\pdfoutput=1
\documentclass[11pt]{article}

\usepackage{acl}

\usepackage{times}
\usepackage{latexsym}

\usepackage[T1]{fontenc}
\usepackage[utf8]{inputenc}

\usepackage{microtype}

\usepackage{inconsolata}

\usepackage{graphicx}
\usepackage{acl}
\usepackage[shortlabels]{enumitem}
\usepackage{times}
\usepackage{color,soul} 
\usepackage{booktabs}
\usepackage{enumitem}

\usepackage{latexsym}
\usepackage{hyperref}
\usepackage{multirow}
\usepackage[LAE,T1]{fontenc}
\usepackage{arabtex}
\usepackage{utf8} % Needed by arabtex package
\usepackage[utf8]{inputenc} % Needed by LAE, T1 for inline arabic
\usepackage[LAE,T1]{fontenc}
\usepackage[arabic,french,english]{babel}
\usepackage[raggedrightboxes]{ragged2e}
\usepackage{todonotes}
\usepackage{array}
\usepackage{lscape}
\usepackage{natbib}
\usepackage{longtable}
\usepackage{xcolor, colortbl}
\usepackage{multirow}
\usepackage{float}
\usepackage{makecell}
\usepackage{inconsolata}
\usepackage{arydshln}
\usepackage{array}
\usepackage{soul}
\usepackage{times}
\usepackage{latexsym}

\usepackage{multirow}
\usepackage{microtype}
\usepackage{hyperref}
\usepackage{microtype}
\usepackage{rotating}
\usepackage{amsmath}
\usepackage{graphicx}   % for including graphics
\usepackage{subfigure}  % for subfigures
\usepackage[utf8]{inputenc}
\usepackage[T1]{fontenc}

\usepackage{inconsolata}

\usepackage{graphicx}
\usepackage{subcaption}
\usepackage{multirow}
\usepackage{tabularx}
\usepackage{framed}
\usepackage{authblk}

\usepackage{colortbl} % for coloring tables
\definecolor{lightgray}{rgb}{0.83, 0.83, 0.83}

\definecolor{lightgreen}{rgb}{0.8, 1, 0.9}  % Adjust RGB values for very light green

\setcode{utf8}

\title{uDistil-Whisper: Label-Free Data Filtering for Knowledge Distillation in Low-Data Regimes}

\author[1]{Abdul Waheed\thanks{\texttt{abdulw@cs.cmu.edu}}}
\author[2]{Karima Kadaoui\thanks{\texttt{karima.kadaoui@mbzuai.ac.ae}}}
\author[1,2]{Bhiksha Raj\thanks{\texttt{bhiksha@cs.cmu.edu}}}
\author[3,2,4]{Muhammad Abdul-Mageed\thanks{\texttt{muhammad.mageed@ubc.ca}}}

\affil[1]{Carnegie Mellon University}
\affil[2]{MBZUAI}
\affil[3]{The University of British Columbia}
\affil[4]{Invertible AI}
\begin{document}
\maketitle
\section*{~~~~~~~~~~~~~~~~~~~~~~~~~~~~~Abstract}
Recent work on distilling Whisper's knowledge into small models using pseudo-labels shows promising performance while reducing the size by up to 50\%. This results in small, efficient, and dedicated models. However, a critical step of distillation using pseudo-labels involves filtering high-quality predictions and using only those during training. This step requires ground truth labels to compare with and filter low-quality examples, making the process dependent on human labels. Additionally, the distillation process requires a large amount of data thereby limiting its applicability in low-resource settings. To address this, we propose a distillation framework that does not require \textit{any} labeled data. Through experimentation, we show that our best-distilled models outperform the teacher model by 5-7 WER points and are on par with or outperform similar supervised data filtering setups. When scaling the data, our models significantly outperform all zero-shot and supervised models. Our models are also 25-50\% more compute- and memory-efficient while maintaining performance equal to or better than that of the teacher model. For more details about our models, dataset, and other resources, please visit our GitHub page: \url{https://github.com/UBC-NLP/uDistilWhisper}.

\section{Introduction}
% \begin{itemize}
%     \item Research gap - What the issue in ~\cite{gandhi2023distil} and Waheed et. al. (our work)
%     \item Why it's important to address that?
%     \item How we are addressing it?
%     \item Briefly describe the motivation and results. 
%     \item Summary the contribution followed by paper outline. 
% \end{itemize}

To democratize automatic speech recognition (ASR), significant attention has been given to multilingual models~\cite{pratap2023scaling, whisper, sm4t, usm}. These powerful systems, thanks to their massive number of parameters and training data, can simultaneously transcribe in hundreds of languages. That being said, low-resource languages tend to trail behind in performance compared to high-resource counterparts such as English~\cite{whisper}. For instance, while OpenAI's latest ASR model Whisper-large-v3 shows the word error rate (WER) in the single digits on English and Spanish, a much lower performance is shown for East Asian and African languages\footnote{\url{https://github.com/openai/whisper/discussions/1762}}~\cite{talafha2023n, talafha2024casablanca}. While the performance reported on these multilingual models is not as low for Arabic, it is important to highlight that it does not fully and accurately represent the diversity of the different varieties that characterize the Arabic language family~\cite{abdul-mageed-etal-2020-toward}. In addition to the disparity in performance, multilingual models are extremely compute-intensive and are thus not equally accessible to everyone. In order to alleviate this issue, knowledge distillation has been the method of choice for multiple works~\cite{hinton2015distillingknowledgeneuralnetwork, yang2023knowledge, sanh2020distilbert, frantar2022gptq}. 

Knowledge distillation has proven to be highly effective in reducing model size, thereby lowering compute and memory requirements while maintaining performance comparable to larger teacher models. For example, \citet{gandhi2023distilwhisper} and \citet{waheed2024distillar} demonstrate its effectiveness for English and Arabic, respectively. However, a significant limitation in these works is their reliance on ground-truth labels to filter out low-quality pseudo-labels generated by the teacher model—a resource that is often scarce, particularly for low-resource languages. Additionally, \citet{waheed2024distillar} shows that training on unfiltered data leads to suboptimal performance. This reliance on labeled data underscores the need for an unsupervised approach to data filtering in knowledge distillation, enabling distilled models to perform better without depending on any ground-truth labels.

% The current gaps that we're aiming to bridge are: (1) the cumbersome size of multilingual ASR models hindering potential progress to their du 
In this work, we propose frameworks to distill knowledge from teacher models that do not rely on filtered pseudo-labels. Our method achieves a similar performance to supervised distillation and outperforms pseudo-label distillation without filtering. Our contributions are the following:

\begin{itemize}
    \item We explore various unsupervised methods to filter out low-quality pseudo-labels, eliminating the need of labeled data for distillation. 
    \item We assess the performance of our distilled models, which effectively outperform the teacher model by 5–7 WER points, and extend the existing setup to include new datasets in both Arabic and Swahili.
    \item We analyze the effectiveness of our best metrics in detecting low-quality pseudo-labeled examples, achieving an AUC ranging between $0.77$ and $0.82$ for examples with WER > 20.
\end{itemize}

% We explore various unsupervised methods to filter out low-quality pseudo-labels, eliminating the necessity of labeled data for distillation.
% We assess the performance of our distilled models across diverse settings and expand our evaluation to include the SADA dataset.
% We analyze the effectiveness of these metrics in detecting low-quality pseudo-labeled examples.

We organize the paper as follows: we present a review of the existing literature in Section~\ref{sec:related_work} before introducing knowledge distillation along with our methodology in Section~\ref{sec:methodology}. Section~\ref{sec:experiments} contains particulars of our experiments and Section~\ref{sec:results} delves into the results we obtain. We discuss these results in Section~\ref{sec:results}, conclude in Section~\ref{sec:conclusion}, and outline our work's limitations in Section~\ref{sec:limitations}.

% - Research gap: low-resource languages don't get enough attention -> only available in massive multilingual models -> models too compute-hungry to be widely accessible. Solution: knowledge distillation. Problem: requires large amounts of high-quality pseudo-labels to effectively train student models. Filtering high-quality labels require GT and not all languages have that option -> need for unsupervised kd. 

% - We select n languages in varying levels of resource availability, through different innovative methods of distillation, we manage to achieve a student performance similar to filtered pseudo labels performance and superior to pseudo-labels without filtering.

% Adding related_work.tex input
\section{Related Work}\label{sec:related_work}

% \begin{itemize}
%     \item Unsupervised KD in ASR 
%     \item Monolingual/Multilingual UKD 
% \end{itemize}

% \textcolor{red}{AbdulWaheed2AbdulWaheed: Taken from ACL work. Need to review, revise, and re-organize.}
% AbdulWaheed: Re-structured related work
Scaling speech foundation models in both data and model size has led to systems capable of handling a wide range of speech tasks~\cite{pratap2023scaling,usm,whisper,sm4t}. These models effectively transcend language barriers, demonstrating robust performance across numerous languages, including many low-resource ones~\cite{hsu2024metawhisperspeechbasedmetaiclasr, yeo2024visualspeechrecognitionlanguages}. However, these models face two key limitations: (1) their large size~\cite{whisper, sm4t}, which results in high computational costs, and (2) the under-representation of low-resource languages in training, leading to suboptimal performance on these languages~\cite{whisper,talafha2023n}. To address the first issue various efficient decoding methods~\cite{faster_whisper, segalfeldman2024whispermedusasearmultihead, malard2023bigmodelhardaudios, leviathan2023fastinferencetransformersspeculative} were proposed. However, resulting models do not necessarily enjoy a reduced memory footprint.

To address the efficiency issue, knowledge distillation~\cite{hinton2015distillingknowledgeneuralnetwork, Gou_2021} is proven to be very effective. Different approaches for knowledge distillation have been explored such as Patient Knowledge Distillation (PKD)~\cite{sun2019patient}, T-S learning~\cite{ts-learning}, Cross-Modal Hashing (CMH)~\cite{hu2020creating}, Joint Unsupervised Domain Adaptation with KD~\cite{zhang-etal-2021-matching} and Data-Free KD~\cite{data-free-kd} among others in the NLP and CV fields alike. Knowledge distillation has also been applied to speech recognition: ~\citet{shao2023whisperkdq} shrink a Whisper~\cite{whisper} model tangibly ($\sim80\%$ of its original size), all while improving its performance; ~\citet{distilhubert} reduce a Hubert model to 75\% of its initial size through layer-wise distillation without a significant drop in performance.

In addition, knowledge distillation has found successful applications in speech-to-text tasks as well, ~\cite{Nayem2023, hentschel2024decodingparalleleffectiveknowledge, tian22_interspeech} effectively reducing the memory and compute requirements. More recently, it has been used to distill multilingual models from Whisper~\cite{ferraz2024multilingualdistilwhisperefficientdistillation} and strong monolingual models~\cite{gandhi2023distilwhisper}, as well as in low-resource settings. An example of the latter is~\citet{waheed2024distillar} who investigate pseudo-label distillation methods~\cite{gandhi2023distilwhisper} across various Modern Standard Arabic (MSA) and dialectal Arabic datasets. Their exploration demonstrate the efficacy of distillation in enhancing both efficiency and performance, showcasing that smaller dedicated models can outperform larger multilingual ones. While the pseudo-labeling approach has proven effective across diverse languages~\cite{gandhi2023distilwhisper, waheed2024distillar}, it traditionally requires labeled data for filtering low-quality pseudo-labeled examples. In this work we address this limitation by introducing an \textit{unsupervised} framework based on data filtering methods, thereby eliminating the dependency on labeled data altogether.

\section{Methodology}\label{sec:methodology}

\subsection{Knowledge Distillation}\label{knowledge-distillation}
% \KK{i removed the equations since they're no longer very relevant to this work}
% \textcolor{red}{Abdul2Himself: Mathematical notations are borrowed (and appropriately cited) from DistilPWhisper paper.}\\
Knowledge distillation is a framework through which a small student model learns the behavior of a bigger teacher model~\cite{hinton2015distillingknowledgeneuralnetwork, sanh2020distilbert, kim-rush-2016-sequence}. \citet{gandhi2023distilwhisper} introduce knowledge distillation via pseudo-labeling, which generates (English) predictions from a teacher Whisper model and filters them through a WER threshold to only keep the most accurate ones. These pseudo-labels are then used to train a smaller student model. \citet{waheed2024distillar} build on this approach and show that a Whisper student model beats its teacher model in its average performance in an Arabic multi-dialectal setting. We follow this standard student-teacher framework to distill Whisper into small yet powerful models. The objective of the distillation process can be stated as:
\begin{align}\label{}
    \mathcal{L}_{KD} = \alpha_{KL} \mathcal{L}_{KL} + \alpha_{PL} \mathcal{L}_{PL}
\end{align}
where:
\begin{itemize}
    \item \( \mathcal{L}_{KL} \) is the Kullback-Leibler (KL) divergence loss, encouraging the student model to match the teacher's probability distribution.
    \item \( \mathcal{L}_{PL} \) trains the student using pseudo-labels as ground truth.
\end{itemize}

The coefficients \( \alpha_{KL} \) (0.8) and \( \alpha_{PL} \) (1.0) balance the contributions of each loss in the overall distillation loss \( \mathcal{L}_{KD} \).

\subsection{Label-Free Data Filtering}\label{subsec:data_filtering}

Previous data filtering methods, such as those used by~\citet{gandhi2023distilwhisper} and~\citet{waheed2024distillar}, rely on ground-truth labels by computing the WER between teacher-generated pseudo-labels and reference transcripts, discarding examples with high error rates. However, this dependence on labeled data limits their use in low-resource languages where such labels are scarce.

Our approach introduces label-free filtering methods that assess pseudo-label quality using the teacher model’s logits, synthetic speech, proxy models, and multimodal embeddings. Each method is detailed below.

\subsubsection{Proxy Models}
We use a pre-trained ASR model, SeamlessM4T-large-v2~\cite{sm4t}, as a proxy to generate reference transcripts for the input speech. Then, the quality of pseudo-labels generated by the teacher model is evaluated by calculating the WER between the proxy's and the teacher's outputs, referred to as pWER. Lower pWER values indicate higher agreement and examples exceeding a defined pWER threshold are discarded.

\subsubsection{Uncertainty Quantification}
We leverage uncertainty in the teacher's output to filter low-quality examples, using two common metrics:

\noindent\textbf{Entropy.} Entropy measures uncertainty in the teacher's predicted probability distribution. High entropy suggests low confidence. It is calculated as:
\begin{equation}
H = -\sum_{i=1}^{N} p_i \log_2(p_i)
\end{equation}
where $p_i$ is the predicted probability for the $i^{\text{th}}$ word.

\noindent\textbf{Geometric Mean of Confidence Scores.} The geometric mean of the confidence scores, representing the probability of each decoded token in the teacher-generated pseudo-label, is used to assess the overall confidence in the pseudo-label. It is calculated as:
\begin{equation}
G = \sqrt[N]{\prod_{i=1}^{N} c_i}
\end{equation}
where $c_i$ is the confidence score for each word. Low entropy and high geometric mean indicate better pseudo-labels.

\subsubsection{Negative Log-Likelihood}
We compute the negative log-likelihood (NLL) using AceGPT-7B~\cite{huang2024acegpt} to assess the quality of pseudo-labels. Lower NLL suggests higher pseudo-label accuracy. It is calculated as:
\begin{equation}
NLL = -\sum_{t=1}^{T} \log(p(y_t | y_{1...t-1}))
\end{equation}
where $p(y_t | y_{1...t-1})$ is the probability assigned to the $t^{\text{th}}$ word.

\subsubsection{Multimodal Embeddings}
We use SONAR~\cite{duquenne2023sonar} to generate embeddings for input speech and pseudo-labels. The dot product is used to compute similarity, with higher scores indicating better alignment and pseudo-label quality.

\subsubsection{Perceptual Evaluation of Speech Quality (PESQ)}
Synthetic speech is generated from pseudo-labels using XTTS-v2\footnote{\url{https://huggingface.co/coqui/XTTS-v2}} and compared with input speech. Similarity is assessed via PESQ, where high scores indicate higher label accuracy.

\subsection{Training Data}
We follow~\citet{waheed2024distillar} for the data mixture to train our models. As a result, we randomly select segments from a diverse set of datasets, including MGB2~\cite{ali2019mgb2}, MGB3~\cite{ali2017speech}, FLEURS~\cite{conneau2022fleurs}, CommonVoice 15.0~\cite{ardila2020common}, QASR~\cite{mubarak2021qasr}, the Arabic Speech Corpus~\cite{halabi2016modern}, and the Massive Arabic Speech Corpus (MASC)~\cite{e1qb-jv46-21}. Specifically, we sample 100K and 500K segments, equivalent to approximately 100 and 500 hours of pseudo-labeled speech data, respectively. Our dataset compilation strictly includes the train splits only from each source. We filter out roughly 27\% low quality examples from our training data using the metrics described in Section~\ref{sec:methodology}.

% \mam{Can you put some details about this filtering in the appendix?} 
% \noindent\textbf{Swahili Data.}

% Moving this part into experiments 
\subsection{Teacher and Student Models}
In line with previous work~\cite{gandhi2023distilwhisper, waheed2024distillar}, we utilize the \textit{whisper-large-v2} checkpoint\footnote{\url{https://huggingface.co/openai/whisper-large-v2}} for both pseudo-labeling and as the teacher model during training. We train two variants of students, differing in the number of layers removed from the teacher model. Following~\citet{gandhi2023distilwhisper} and~\citet{waheed2024distillar}, we initialize the student models with maximally spaced layers in the encoder and decoder block of the teacher model. When training on 100K segments, we refer to the models with 16-16 and 32-16 encoder-decoder blocks as \textit{UDW-16-16}\footnote{Following the format from \citet{waheed2024distillar}, UDW refers to Unsupervised Distill Whisper.} and  \textit{UDW-32-16}, respectively. Similarly, we refer to the models trained on 500K segments as \textit{UDW-16-16++} and \textit{UDW-32-16++}.

% Further details on our distilled models are presented in Table~\ref{distilled-models}.
% \input{sections/tables/distilled-models}

\section{Experiments}\label{sec:experiments}
For a fair comparison and analysis, we keep our experimental setup quite identical to~\citet{waheed2024distillar}. Apart from evaluating our models on five standard benchmark datasets, we extend the evaluation to include the SADA~\cite{10446243} and Casablanca\footnote{The In-House data in~\citet{waheed2024distillar}, reported in Table \ref{tab:results_main}, is an earlier subset of the Casablanca dataset, used prior to the official release of the larger version.}~\cite{talafha2024casablanca} datasets. We describe all the used datasets below.

\subsection{Evaluation Dataset}\label{subsec:dataset}
% \begin{itemize}
%     \item How evaluation data is different from training data. 
%     \item Add PCA features of Whisper to demonstrate distribution drift. Include SADA data as well. 
%     \item 
% \end{itemize}

\noindent\textbf{FLEURS}. The Few-shot Learning Evaluation of Universal Representations of Speech (FLEURS)~\cite{conneau2022fleurs} is a multilingual speech dataset with 102 languages, each with ~12 hours of speech. It supports tasks like ASR, language identification, and translation. We use the Arabic subset, which features MSA speech with Egyptian accents.

\noindent\textbf{Common Voice}. Common Voice (CV)~\cite{ardila2020common} is a volunteer-driven multilingual dataset with 124 languages and 31,176 hours of speech. The Arabic subset includes 156 hours of primarily MSA speech.

\noindent\textbf{Multi-Genre Broadcast}. The MGB dataset~\cite{ali2019mgb2, ali2017speech, 9003960} consists of MGB2 (1,200 hours of MSA-dominant Aljazeera Arabic speech), MGB3 (six hours of Egyptian dialectal speech), and MGB5 (14 hours of Moroccan dialectal speech).

\noindent\textbf{SADA}. The Saudi Audio Dataset for Arabic (SADA)~\cite{10446243} includes 668 hours of Arabic speech (435 labeled), featuring Saudi dialects, MSA, and speech from the Levant, Yemen, and Egypt. Statistics are provided in Appendix~\ref{appendix:dataset}~\ref{appendix:sada_dataset}.

\noindent\textbf{Casablanca}. Casablanca~\cite{talafha2024casablanca} is a dialectal Arabic dataset that contains $48$ hours of speech from various TV shows. It covers eight different dialects and is annotated by native speakers. The results on this dataset are separately reported in Table \ref{tab:casablanca_results_wer}.

% \input{tables/sada-stats}. 
% \mam{Move this table to appendix if you need the space.}  Done
% # \KK{add table about dataset distribution}

\subsection{Baselines}
% \AW{Abdul: Text Copied from ACL work}
% \AW{Update: ACL work revised but needs to be reviewed.}

We employ several monolingual and multilingual  speech recognition models on different Arabic varieties, including standard and accented MSA, as well as various dialects. The models are grouped as follows:

\subsubsection{Supervised Models and Commercial Systems.} We evaluate three baselines, namely Wav2Vec2-XLS-R \cite{conneau2020unsupervised, babu2021xlsr} trained on CV8.0, HuBERT \cite{hsu2021hubert} trained on both MGB-3~\cite{ali2017speech} and a 5.5-hour Egyptian Arabic corpus, and a fine-tuned Whisper-large-v2 model trained on CV11.0 and MGB-2. We use these models off-the-shelf from their respective checkpoints. 

% We also report the results of Amazon Transcribe\footnote{\url{https://aws.amazon.com/transcribe/} on five dialects evaluate in~\citet{waheed2024distillar}. 

\subsubsection{Zero-Shot Models}

\noindent\textbf{Whisper.} Whisper~\cite{whisper} is a versatile multilingual speech model designed for both speech recognition and translation, including Arabic. We assess four Whisper variants: \textit{whisper-small} (W-S), \textit{whisper-medium} (W-M), \textit{whisper-large-v2} (W-L-v2), and \textit{whisper-large-v3} (W-L-v3), using their default decoding parameters with a maximum sequence length of 225 tokens.

\noindent\textbf{SeamlessM4T.} SeamlessM4T models excel at generating high-quality transcripts across languages~\cite{sm4t}, but evaluations on them typically focus on English. To bridge this gap, we test three of its versions -- \textit{seamless-m4t-medium} (SM4T-M), \textit{seamless-m4t-v1-large} (SM4T-L-v1), and \textit{seamless-m4t-v2-large} (SM4T-L-v2) -- under a zero-shot setting on Arabic ASR, using the default parameters from the model's inference pipeline.

% \footnote{\url{https://huggingface.co/facebook/seamless-m4t-v2-large}}.

% \noindent\textbf{Commercial Systems.}
% % \textcolor{red}{AbdulWaheed2Karima: Can you please add a little bit about AmazonASR? Thanks.} 

\subsubsection{Distilled Models}\label{distilled-models}
To compare our results with an equivalent supervised setup, two distilled model variants of different sizes are evaluated. We extend the evaluation to SADA's \textit{test} and \textit{valid} splits, using 16-16 and 32-16 models trained with a WER threshold of 80\% on both 100K and 500K segments. Additionally, we include models trained without filtering as a lower bound. For a fair comparison with supervised data filtering, we follow an experimental setup identical to prior work~\cite{waheed2024distillar}.

% As described in Section \ref{knowledge-distillation}, we distill \textit{whisper-large-v2} into seven different student models (see Table~\ref{distilled-models}). We provide more details about the teacher and student models and distillation data here. 

% \AW{Abdul: Taken from ACL work}
\subsection{Experimental Setup}
% In addition to automatic evaluation, we also conduct human evaluation of the errors that these models are making. 
During evaluations, we adhere to the default decoding parameters specified by the original model implementations unless specified otherwise. Throughout our experiments, a maximum sequence length of 225 is maintained and WER and CER are used as the primary metrics for the quantitative evaluation of the models.

For distillation, roughly 27\% of the top segments are selected based on the metrics, to match the amount of data used in a supervised setup when $\lambda=80\%$. We do not conduct hyperparameter search due to computational constraints, and instead apply the configuration detailed in~\newcite{gandhi2023distilwhisper}. The distillation process and its key parameters are detailed in Table~\ref{training-parameters} in Appendix~\ref{appendix:experiments}.

% \noindent\textbf{Text Preprocessing.}\label{text-preprocessing}
% Arabic text presents unique challenges for ASR due to its variable use of diacritics and letter forms (e.g. {\small\<أ>} vs {\small\< ا>} ). Such inconsistencies can affect evaluation accuracy since phonetically correct transcriptions might be marked incorrect over minor lexical discrepancies. To mitigate these issues, we adopt text normalization protocols from~\citet{talafha2023nshot, chowdhury2021model}, which include removing diacritics, excluding Latin characters to avoid code-switching complications, converting Arabic digits (i.e. \<١>, \<٢>, \<٣>) to their numerals (1, 2, 3), and normalizing all variants of \textit{alef} to a standard form without \textit{hamza}.

\section{Results and Discussion}\label{sec:results}
% \AW{Can we add some plots for the results and put tables in the Appendix to enhance readability?}
% Please add the following required packages to your document preamble:
% \usepackage{multirow}
% for cdashline \cdashline{1-1}[1pt/2pt]{}
\newcolumntype{G}{>{\columncolor{lightgreen}}c}

\begin{table*}[h!]
\centering
\Large

\resizebox{1.\linewidth}{!}{% Adjust the table size to fit the width of the text

\begin{tabular}{lcccccccccccccc}
% Please add the following required packages to your document preamble:
% \usepackage{multirow}
\toprule
 \multirow{2}{*}{Model} & \multirow{2}{*}{Size}  & \multirow{2}{*}{CV15.0} & \multirow{2}{*}{MGB2} & \multirow{2}{*}{MGB3} & \multirow{2}{*}{MGB5} & \multirow{2}{*}{Fleurs} & \multicolumn{5}{c}{In-house Data} & \multirow{2}{*}{SADA}                                     \\
                                      &          &                       &                       &                       &                         & & ALG          & JOR         & PAL         & UAE          & YEM          \\ \midrule
                                      \textbf{Baselines}\\
                                      \midrule
  Amazon & -- & -- & -- & -- & -- & -- & 83.6 & 45.5 & 52.4 & 58.8 & \underline{64.7} & -- \\
\noalign{\vskip 0.3ex} % extra space above the dashline
\hdashline
\noalign{\vskip 0.6ex} % extra space below the dashline

 XLS-R & 0.96 & 89.7 & 97.6 & 98.7 & 99.5 & 94.9 & 99.7 & 99.1 & 99.1 & 99.4 & 99.5 & 99.5 \\

 HuBERT & 0.31 & 55.2 & 49.6 & \textbf{25.2} & 92.4 & 34.9 & 96.8 & 65.2 & 73.8 & 83.0 & 90.5 & 75.6 \\

 W-FT & 1.5 & 35.8 & \textbf{15.3} & 48.9 & 101.4 & 9.8 & 115.5 & 67.8 & 69.6 & 105.9 & 107.1 & 92.3 \\ \midrule

 MMS-all & 1.0 & 106.4 & 39.3 & 75.3 & 89.7 & 23.8 & 100.2 & 89.8 & 99.9 & 100.1 & 100.2 & 	77.6  \\
\noalign{\vskip 0.3ex} % extra space above the dashline
\hdashline
\noalign{\vskip 0.6ex} % extra space below the dashline

 SM4T-M & 1.2 & 16.3 & 19.5 & 41.4 & 83.8 & \underline{8.7} & 81.1 & 46.3 & 55.2 & 59.8 & 68.9 & 65.2 \\

 SM4T-L-v1 & 2.3 & 19.8 & 21.8 & 44.4 & 89.9 &  11.1 & 87.9 & 50.7 & 57.5 &  61.8 & 72.2 & 64.9 \\

 SM4T-L-v2 & 2.3 & \textbf{11.3} & 17.3 & 36.2 & 89.1 & \textbf{7.6} & 92.1 & \textbf{41.5} & 49.5 & 55.9 & 69.7 & 65.0 \\
\noalign{\vskip 0.3ex} % extra space above the dashline
\hdashline
\noalign{\vskip 0.6ex} % extra space below the dashline

 W-S & 0.24 & 40.3 & 46.8 & 81.4 & 226.5 & 28.2 & 130.7 & 68.6 & 73.8 & 97.8 & 107.1 & 139.9 \\

 W-M & 0.77 & 29.8 & 33.1 & 64.3 & 127.7 & 16.4 & 103.7 & 50.5 & 58.7 & 82.5 & 86.8 &  99.3 \\

 W-L-v2 & 1.5 & 19.6 & 26.5 & 53.0 & 99.2 & 11.4 & 106.4 & 42.3 & 51.1 & 63.8 & 77.3 & 69.8 \\

 W-L-v3 & 1.5 & 15.8 & \underline{15.9} & \underline{35.7} & 79.8 & 9.7 & 101.9 & 43.6 & 53.4 & 63.4 & 76.1 & 67.2 \\ 
 \midrule

% % commenting this -> uncomment it if it's required for comparison
% & DW-8-8 & 0.44 & 32.7 & 39.6 & 64.9 & 89.7 & 29.8 & 91.4 & 66.2 & 73.2 & 78.0 & 82.9 & 80.7 \\

 DW-16-16 & 0.80 & 22.1 & 26.0 & 50.5 & 82.4 & 18.8 & 83.0 & 50.4 & 61.0 & 64.6 & 72.7 & 66.7 \\

 DW-32-16 & 1.12 & 18.8 & 21.1 & 43.8 & \underline{78.9} & 14.2 & 79.5 & 44.4 & 55.0 & 58.1 & 68.5 & 66.9 \\

% % commenting this -> uncomment it if it's required for comparison
% & DW-16-32 & 1.22 & 21.5 & 25.0 & 49.1 & 83.0 & 18.4 & 84.3 & 49.8 & 60.3 & 64.4 & 73.8 & 61.4 \\ 
\noalign{\vskip 0.3ex} % extra space above the dashline
\hdashline
\noalign{\vskip 0.6ex} % extra space below the dashline

 DW-16-16++ & 0.80 & 19.2 & 23.0 & 47.2 & 79.0 & 15.0 & 79.0 & 46.7 & 56.4 & 60.4 & 69.1 & 69.9  \\
 DW-32-16++ & 1.12 & 17.1 & 19.7 & 40.7 & \textbf{76.6} & 11.1 & 74.6 & \underline{41.6} & 51.4 & \underline{53.5} & \textbf{63.5} & 60.3 \\
\noalign{\vskip 0.3ex} % extra space above the dashline
\hdashline
\noalign{\vskip 0.6ex} % extra space below the dashline

 No-filter \\
 $\quad-$ DW-16-16 & 0.80 & 22.8 & 26.1 & 54.1 & 95.1 & 17.6 & 93.4 & 51.8 & 64.7 & 68.0 & 78.3 & 78.4 \\ 
 $\quad-$ DW-32-16 & 1.12 & 21.2 & 22.8 & 51.3 & 90.5 & 14.9 & 87.3 & 47.6 & 57.1 & 63.9 & 73.0 & 70.7\\
\midrule
\textbf{Ours}\\
\midrule

% \rowcolor{lightgreen}
 UDW-16-16 & 0.80 \\
% \rowcolor{lightgreen}
 $\quad-$ nll &  & 23.6 & 26.4 & 54.4 & 92.3 & 18.5 & 82.4 & 63.3 & 53.0 & 72.0 & 93.4
 & 84.3 \\ 
% \rowcolor{lightgreen}
 $\quad-$ pesq &  & 24.3 & 28.3 & 54.2 & 93.7 & 19.7 & 81.5 & 64.9 & 52.9 & 69.7 & 87.0 & 79.7 \\ 
% \rowcolor{lightgreen}
 $\quad-$ entropy & & 23.8 & 27.4 & 57.3 & 94.8 & 18.2 & 90.1 & 54.8 & 64.3 & 73.7 & 96.4 & 92.0  \\

% \rowcolor{lightgreen}
 $\quad-$ conf & & 23.5 & 27.2 & 52.7 & 87.8 & 17.7 & 89.2 & 53.1 & 63.4 & 69.0 & 79.1 & 78.6  \\

% \rowcolor{lightgreen}
 $\quad-$ proxy & & 22.5 & 25.5 & 49.4 & 84.8 & 17.6 & \underline{74.5} & 61.1 & 50.5 & 65.4 & 84.1 & 68.6 \\ 
% \rowcolor{lightgreen}
 $\quad-$ sonar & & 24.1 & 28.3 & 55.1 & 85.6 & 20.4 & 86.9 & 56.4 & 65.4 & 70.7 & 76.5 & 74.9 \\
\noalign{\vskip 0.3ex} % extra space above the dashline
\hdashline
\noalign{\vskip 0.6ex} % extra space below the dashline

% \rowcolor{lightgreen}
 UDW-32-16 & 1.12 \\
% \rowcolor{lightgreen}
 $\quad-$ nll & & 18.9 & 22.6 & 48.1 & 94.1 & 13.3 & 75.4 & 55.2 & \underline{44.7} & 63.7 & 83.8
 & 73.4 \\ 
% \rowcolor{lightgreen}
 $\quad-$ pesq & & 21.7 & 24.2 & 49.0 & 87.0 & 16.1 & 88.8 & 47.5 & 58.5 & 67.0 & 75.5 & 73.3 \\

% \rowcolor{lightgreen}
 $\quad-$ entropy & & 19.1 & 21.8 & 46.9 & 87.1 & 13.3 & 91.0 & 46.7 & 58.2 & 64.9 & 76.4 & 68.3  \\

% \rowcolor{lightgreen}
 $\quad-$ conf & & 20.3 & 22.8 & 46.6 & 83.4 & 14.5 & \textbf{73.4} & 57.3 & 61.1 & \textbf{47.5} & 85.7 & 66.2  \\

% \rowcolor{lightgreen}
 $\quad-$ proxy & & 19.0 & 21.5 & 44.3 & 82.6 & 14.2 & 80.4 & 45.5 & 56.0 & 61.6 & 69.5 & 64.6 \\ 
% \rowcolor{lightgreen}
 $\quad-$ sonar & & 17.8 & 21.1 & 45.3 & 80.4 & 13.1 & 79.0 & 44.3 & 54.3 & 58.8 & 66.8
 & 58.7 \\
\noalign{\vskip 0.3ex} % extra space above the dashline
\hdashline
\noalign{\vskip 0.6ex} % extra space below the dashline

% \rowcolor{lightgreen}

% \rowcolor{lightgreen}
 UDW-16-16++ & 0.80 \\
% \rowcolor{lightgreen}
 $\quad-$ proxy & & \underline{15.5} & 23.3 & 46.8 & 84.6 & 14.2 & 82.9 & 46.4 & 57.0 & 61.1 & 70.8 & \underline{60.2} \\ 
% \rowcolor{lightgreen}
 $\quad-$ sonar & & 18.9 & 22.8 & 47.2 & 82.0 & 14.9 & 81.5 & 47.7 & 57.4 & 62.1 & 69.2 & 59.4 \\
% \rowcolor{lightgreen}
 UDW-32-16-++ & 1.12 \\
% \rowcolor{lightgreen}
 $\quad-$ proxy & & 18.1 & 20.8 & 43.1 & 82.3 & 12.6 & 80.8 & 43.1 & 54.4 & 57.7 & 68.5 & \textbf{55.1}  \\ 
% \rowcolor{lightgreen}
 $\quad-$ sonar & & 17.0 & 21.9 & 44.1 & 79.8 & 12.8 & 78.2 & 54.2 & \textbf{44.4} & 58.2 & 66.6 & 61.6 \\

\bottomrule 
\end{tabular}
}
\caption{ \label{tab:results_main}
WER ($\downarrow$) after normalization and removing diacritics. All baseline distilled models (dw-) are trained with a filtering threshold of 80 if not specified. Best results are shown in \textbf{bold}. Second best results are \underline{underlined}. We report the score on the test split of each dataset.
Abbreviations. \textbf{W} - Whisper, \textbf{FT} - Finetuned, \textbf{M} - Medium, \textbf{L} - Large, \textbf{S} - Small, \textbf{DW} - Distill Whisper, \textbf{UDW} - Unsupervised Distill Whisper, \textbf{nll} - negative log likelihood, \textbf{conf} - confidence score.
 % \KKcomment{leave only baselines and our best model, move other proposed methods to a different table without baselines}
 }
\end{table*}

We report the baseline results from~\citet{waheed2024distillar} and evaluate all models on SADA's test and validation splits. Our main results can be found in Table~\ref{tab:results_main}, and the average across different evaluation setups in Table~\ref{tab:results_average}. Results on the top five varieties\footnote{We refer labels in SADA as varieties as not all labels are necessarily dialects.} of SADA are provided in Table~\ref{tab:sada_top_five_dialects_results}. The reported results are computed on normalized text that includes removing diacritics. Results on the text with no normalization can be found in Appendix~\ref{appendix:results}.

% ~\citet{waheed2024distillar} show that their supervised models perform well on in-distribution evaluation. However, their performance degrades considerably under novel and challenging conditions. Our evaluation of these supervised models shows similar results. We find that HuBERT trained on MGB3 outperforms all other supervised baselines across different evaluation setups. 

While larger variants of both SeamlessM4T and Whisper perform well on standard benchmark setups like Common Voice and Fleurs, we find that they perform poorly on dialectal speech. For example, the best zero-shot model (SeamlessM4T-large-v2) has a 7.6\% WER on Fleurs compared to 65.0\% on the SADA test set as shown in Table~\ref{tab:results_main}. This underscores the inadequate evaluation and limited generalization capability of zero-shot models in unseen and challenging settings. The best-distilled model DW-32-16++ outperforms all other models including its teacher (large-v2), the best Whisper version (large-v3), and the best baseline (SeamlessM4T-large-v2).

% We train two versions of teacher models using datasets of size 100K and 500K segments, respectively. Approximately 27\% of the examples were removed from each dataset based on specific filtering metrics, matching the amount of data used in the supervised setup when the filtering threshold is 80.
% Consequently, we consider models trained with this threshold of 80 to represent an upper bound, while models trained on the original, unfiltered pseudo labels serve as our baseline for comparison. \KKcomment{all of the info in this parahraph was already mentioned before the results section, I think we can remove it from here} \mam{Yes, this is repeated content.}
% DW-32-16 Supervised avg - 54.33
% UDW-32-16 Sonar avg - 51.6
% DW-32-15 No filter - 58.86
% DW-16-16 supervised - 57.7
% UDW-16-16 proxy - 58.5
% DW_16-16 no filter - 64.2
When it comes to our distilled models, we observe that the two best filtering measures, \textit{sonar-sim} and \textit{proxy-ref}, are comparable to or surpass the equivalent supervised setups. For instance, DW-32-16 trained on 100K examples with a filtering threshold of 80 achieves a WER of 35.3\% on the benchmark test split, 66.6\% on the SADA test split, and 61.1\% averaged on five novel dialects. By comparison, our best-performing model in a similar configuration, UDW-32-16-sonar, achieves 35.5\% WER on the benchmark test split, 58.7\% on the SADA test, and 60.6\% on five dialects. This reflects close to 3\% improvement in average score compared to the supervised setup (51.6 vs 54.3) and more than a 7\% improvement over the no-filter setup, as detailed in Table~\ref{tab:results_average}.
In addition to that, our experiments show that the smaller student model (UDW-16-16) performs better with the \textit{proxy-ref} method. More specifically, when trained on 100K examples, it matches the results of supervised setups (58.5\% vs 57.6\%) and shows over 6\% improvement compared to setups where pseudo-labels are not filtered (64.2\%).

\begin{table}[h!]
\centering
\renewcommand{\arraystretch}{0.8} % Adjust row height
% \setlength{\tabcolsep}{4pt}
% \small 
% \renewcommand{\arraystretch}{1.1}   %to squeeze table content low value means squeeze more
\resizebox{1.\linewidth}{!}{% reduce the text size, lower means smaller text size

\begin{tabular}{lccccc}
\toprule
\multirow{2}{*}{Model} & \multicolumn{2}{c}{Bench} & \multicolumn{2}{c}{SADA2022} & \multirow{2}{*}{IH} \\
                       & Test          & Valid         & Test         & Valid         &     \\ \midrule
\textbf{Baselines}\\
\midrule
HuBERT & 51.4 & 54.4 & 75.6 & 73.7 & 81.9 \\ 

W-FT & 42.2 & 47.4 & 92.3 & 81.7 & 93.2 \\
\noalign{\vskip 0.3ex} % extra space above the dashline
\hdashline
\noalign{\vskip 0.6ex} % extra space below the dashline
           
SM4T-v1 & 37.4 & 39.5 & 64.9  & 60.3 & 66.0 \\

SM4T-v2 & \underline{32.3} & \underline{34.7}  & 65.0 & 63.2 & 61.7 \\

W-M & 54.3 & 59.6 & 99.3 & 91.9  & 76.4 \\

W-L-v2 & 41.9 & 46.8 & 69.8 & 66.5  & 68.2  \\

W-L-v3 & \textbf{31.4} & \textbf{33.0} & 67.2 & 58.4 & 67.7 \\

\midrule

DW-16-16 & 40.0 & 42.2  & 66.7 & 65.2 & 66.3 \\

DW-32-16 & 35.3 & 37.7  & 66.9  & 64.5 & 61.1 \\

DW-16-16++ & 36.7 & 38.8 & 69.9 & 64.7 & 62.3 \\
DW-32-16++ & 33.0 & 34.9 & 60.3 & 57.1 & \textbf{56.9} \\ 

\noalign{\vskip 0.3ex} % extra space above the dashline
\hdashline
\noalign{\vskip 0.6ex} % extra space below the dashline

No-Filter \\
$\quad-$ DW-16-16 & 43.1 & 46.5 & 78.4 & 72.5 & 71.2  \\ 
$\quad-$ DW-32-16 & 40.1 & 43.6 & 70.7 & 66.0 & 65.8  \\ 

\midrule
\textbf{Ours}\\
\midrule      
UDW-16-16 \\
% $\quad-$ entropy & 0.0 & 0.0 & 0.0 & 0.0 & 0.0  \\ 
% $\quad-$ nll & 0.0 & 0.0 & 0.0 & 0.0 & 0.0 \\ 
% $\quad-$ pesq & 0.0 & 0.0 & 0.0 & 0.0 & 0.0  \\ 
$\quad-$ proxy & 40.0 & 42.3 & 68.6 & 65.4 & 67.1  \\ 
$\quad-$ sonar & 42.7 & 44.9 & 74.9 & 71.2 & 71.2  \\
\noalign{\vskip 0.3ex} % extra space above the dashline
\hdashline
\noalign{\vskip 0.6ex} % extra space below the dashline

UDW-32-16 \\
% $\quad-$ entropy & 0.0 & 0.0 & 0.0 & 0.0 & 0.0   \\ 
% $\quad-$ nll & 0.0 & 0.0 & 0.0 & 0.0 & 0.0  \\ 
% $\quad-$ pesq & 0.0 & 0.0 & 0.0 & 0.0 & 0.0 \\ 
$\quad-$ proxy & 36.3 & 39.1 & 64.6 & 58.7 & 62.6  \\ 
$\quad-$ sonar & 35.5 & 37.8 & \underline{58.7} & 56.4 & 60.6  \\

\noalign{\vskip 0.3ex} % extra space above the dashline
\hdashline
\noalign{\vskip 0.6ex} % extra space below the dashline

UDW-16-16++ \\

$\quad-$ proxy & 36.9 & 41.4 & 64.1 & 60.2 & 63.6  \\ 
$\quad-$ sonar & 37.2 & 39.3 & 61.7 & 59.4 & 63.6  \\
\noalign{\vskip 0.3ex} % extra space above the dashline
\hdashline
\noalign{\vskip 0.6ex} % extra space below the dashline

UDW-32-16++ \\
$\quad-$ proxy & 35.1 & 37.2 & 58.9 & \underline{55.1} & \underline{60.3}  \\ 
$\quad-$ sonar & 35.4 & 37.7 & \textbf{57.9} & \textbf{54.7} & 60.9  \\

\bottomrule
\end{tabular}
}
\caption{ \label{tab:results_average}
Average WER across different evaluation datasets. Bench: CV15.0, FLEURS and the three MGBs. IH: In-House data. Best results are shown in \textbf{bold}. Second best results are \underline{underlined}. WER scores are reported after normalization and removing diacritics.
}
% All baseline distilled models (dw-) are trained with a filtering threshold of 80 if not specified. We report the score on the test split of each dataset.
% Abbreviations. \textbf{W} - Whisper, \textbf{FT} - Finetuned, \textbf{M} - Medium, \textbf{L} - Large, \textbf{S} - Small, \textbf{U} - Unsupervised, \textbf{D} - Distil.
% This part of table caption is redundant.
\end{table}

\noindent\textbf{Scaling the Data.}
% UDW 32-16++ proxy average - 51.4
% UDW 32-16++ sonar average - 51.4
% DW 32-16++ average - 50.0
% DW - 16-16 ++ average - 56.3
% UDW 16-16++ proxy average - 54.7
% UDW 16-16++ sonar average - 53.9
% \KKcomment{I think it would be better if we take the entire scaling data section and just merge it with the previous one, like we mention we trained the UDW models on both the 100K and 500K splits and these are the results for each configuration. Cuz otherwise it looks like we introduce the same thing twice}
% That way results look too dull and kinda static
We scale our dataset from 100K to 500K segments and we train two models, UDW-16-16 and UDW-32-16, employing \textit{proxy-ref} and \textit{sonar-sim} as filtering criteria. As shown in Table~\ref{tab:results_average}, our UDW-32-16++ model, trained with 500K segments, performs comparably with its supervised counterpart DW-32-16++. For instance, UDW-32-16++ with \textit{proxy-ref} filtering achieves a WER of 51.4\%, similar to DW-32-16++'s    50.0\%. Notably, our smaller student model (UDW-16-16++) trained on 500K segments results in superior performance compared to models trained under supervised filtering. For example, UDW-16-16++ with \textit{sonar-sim} gives 53.9\% WER compared to 56.2 for DW-16-16++. These results showcase the efficacy of our data filtering methods.

\noindent\textbf{Generalization to Unseen Dialects.}
% DW-32-16++ avg five dialects- 59.42
% UDW-32-16++ proxy avg five dialects- 58.06
We evaluate our models under various unseen conditions, including five novel dialects that our models have not encountered before. Additionally, we compute category-wise~\footnote{By \textit{categories} we refer to dialectal and non-dialectal labels in the SADA data.} error rates for both the SADA test and validation splits. We present results on the top five categories. As detailed in Table~\ref{tab:sada_top_five_dialects_results}, our best model UDW-32-16++ consistently outperforms all others when averaged across these dialects.

\begin{table}[h!]
\centering
\renewcommand{\arraystretch}{0.9} % Adjust row height
\setlength{\tabcolsep}{4pt} % Adjust column spacing
\resizebox{1.\linewidth}{!}{% Resize table to fit within the text width
\begin{tabular}{lccccc}
\toprule
Model & NJD & MTOS & KHLJ & HJZ & UNK \\
\midrule
\textbf{Baselines}\\
\midrule
W-FT & 106.4 & 77.0 & 117.8 & 84.2 & 139.9 \\
SM4T-v1 & 56.5 & 74.0 & 61.9 & 54.6 & 69.6 \\
SM4T-v2 & 52.6 & 76.2 & 58.2 & 53.3 & 74.7 \\
W-M & 88.2 & 107.9 & 102.8 & 86.7 & 134.0 \\
W-L-v2 & 56.6 & 74.0 & 80.2 & 63.4 & 98.2 \\
W-L-v3 & 53.1 & 73.7 & 70.5 & 61.7 & 89.0 \\
\midrule
DW-16-16 & 58.2 & 74.9 & 65.1 & 58.2 & 68.4 \\
DW-32-16 & 58.7 & 74.9 & 66.7 & 58.7 & 70.1 \\
DW-16-16++ & 58.8 & 81.9 & 62.1 & 57.5 & 77.5 \\
DW-32-16++ & 51.7 & 67.8 & \underline{57.5} & 52.6 & 67.5 \\
\noalign{\vskip 0.3ex} % extra space above the dashline
\hdashline
\noalign{\vskip 0.6ex} % extra space below the dashline

No-Filter & & & & &  \\
$\quad-$ DW-16-16 & 61.6 & 90.4 & 72.9 & 68.5 & 88.8 \\
$\quad-$ DW-32-16 & 58.2 & 82.3 & 67.3 & 60.4 & 76.6 \\
\midrule
\textbf{Ours}\\
\midrule
UDW-16-16 & & & & &  \\
$\quad-$ proxy & 60.2 & 75.6 & 65.7 & 62.1 & 75.6 \\
$\quad-$ sonar & 64.6 & \underline{59.9} & 84.6 & 72.1 & 66.6 \\
\noalign{\vskip 0.3ex} % extra space above the dashline
\hdashline
\noalign{\vskip 0.6ex} % extra space below the dashline

UDW-32-16 & & & & &  \\
$\quad-$ proxy & 53.4 & 75.7 & 60.0 & 51.4 & 72.8 \\
$\quad-$ sonar & \textbf{48.8} & \textbf{52.6} & 67.5 & 58.3 & \textbf{48.9} \\
\noalign{\vskip 0.3ex} % extra space above the dashline
\hdashline
\noalign{\vskip 0.6ex} % extra space below the dashline

UDW-16-16++ & & & & &  \\
$\quad-$ proxy & 53.4 & 73.9 & 61.9 & 53.3 & 68.6 \\
$\quad-$ sonar & 53.7 & 69.4 & 61.4 & 53.7 & 64.6 \\
\noalign{\vskip 0.3ex} % extra space above the dashline
\hdashline
\noalign{\vskip 0.6ex} % extra space below the dashline

UDW-32-16++ & & & & &  \\

$\quad-$ proxy & \underline{48.9} & 66.7 & 60.6 & \textbf{48.7} & 66.0 \\
$\quad-$ sonar & 50.5 & 65.3 & \textbf{57.3} & \underline{49.1} & \underline{61.8} \\
\bottomrule

\end{tabular}
}
\caption{ \label{tab:sada_top_five_dialects_results}
WER results on top five dialects/categories on the test set of the SADA data. NJD: Najdi. MTOS: More than one speaker. KHLJ: Khaleeji. HJZ: Hijazi. UNK: Unknown. Best results are shown in \textbf{bold}. Second best results are \underline{underlined}. WER scores are reported after normalization and removing diacritics.}
\end{table}

\begin{table*}[!ht]
    \centering
    % \resizebox{\columnwidth}{!}{%
    \begin{tabular}{lccccccccc}
    \toprule
    Model & ALG & EGY & JOR & MAU & MOR & PAL & UAE & YEM & AVG\\
    \midrule
    \textbf{Baselines}\\
    \midrule
    SM4T-v2 & 94.3 & \textbf{52.3} & \textbf{39.2} & 88.9 & 91.0 & \textbf{49.0} & \textbf{54.7} & \textbf{62.4} & \textbf{66.5}\\
    W-L-v2 & 89.9 & \underline{58.1} & \underline{43.2} & 108.3 & 101.1 & \underline{51.9} & 60.8 & 81.4 & 74.3 \\
    \midrule
    DW-16-16 & 85.7 & 65.3 & 51.1 & 88.5 & 86.2 & 60.2 & 64.1 & 69.3 & 71.3 \\
    DW-32-16 & 86.6 & 65.1 & 51.0 & 89.2 & 87.1 & 59.9 & 63.0 & 69.8 & 71.5 \\
    No-Filter & & & & & & & & & \\
    $\quad-$ DW-32-16 & 93.2 & 66.6 & 48.1 & 95.3 & 86.0 & 58.3 & 63.8 & 68.8 & 72.5 \\
    \midrule
    \textbf{Ours}\\
    \midrule
    UDW-16-16 & & & & & & & & & \\
    $\quad-$ proxy & 85.9 & 67.1 & 51.0 & 88.0 & 87.4 & 61.3 & 63.5 & 70.8 & 71.9 \\
    $\quad-$ sonar & 88.4 & 69.8 & 57.3 & 91.3 & 90.9 & 65.7 & 68.4 & 72.8 & 75.6 \\
    \noalign{\vskip 0.3ex} % extra space above the dashline
    \hdashline
    \noalign{\vskip 0.6ex} % extra space below the dashline
    UDW-32-16 & & & & & & & & & \\
    $\quad-$ proxy & \underline{85.6} & 61.8 & 46.2 & \underline{87.9} & \textbf{84.5} & 57.8 & 59.1 & 64.8 & \underline{68.5}\\
    $\quad-$ sonar & \textbf{82.1} & 58.8 & 45.5 & \textbf{86.0} & \underline{85.5} & 53.3 & \underline{57.2} & \underline{63.3} & \textbf{66.5}\\
    \bottomrule
    \end{tabular}%
    % }
    \caption{Results on the Casablanca dataset. Best results are shown in bold. Second-best results are underlined. WER (↓) scores are reported after normalization and removing diacritics.
We report the score on the test split of each dataset.}
    \label{tab:casablanca_results_wer}
\end{table*}

Furthermore, UDW-32-16++, demonstrates superior performance over DW-32-16++ in the top five SADA categories. For instance, when using \textit{proxy-ref} as a filtering measure, UDW-32-16++ achieves 58.06\% WER, compared to DW-32-16++'s 59.42\% averaged across top (with most utterances) five categories in the SADA \textit{test} split. This demonstrates our ability to (1) distill smaller models from larger Whisper models, (2) maintain or improve performance, and (3) reduce model size without relying on labeled data.

\noindent\textbf{Effectiveness of Unsupervised Metrics to Filter Low-Quality Pesudo Labels.}
\begin{figure*}[ht]
    \centering
    \subfigure[WER > 80]{\includegraphics[width=0.30\textwidth]{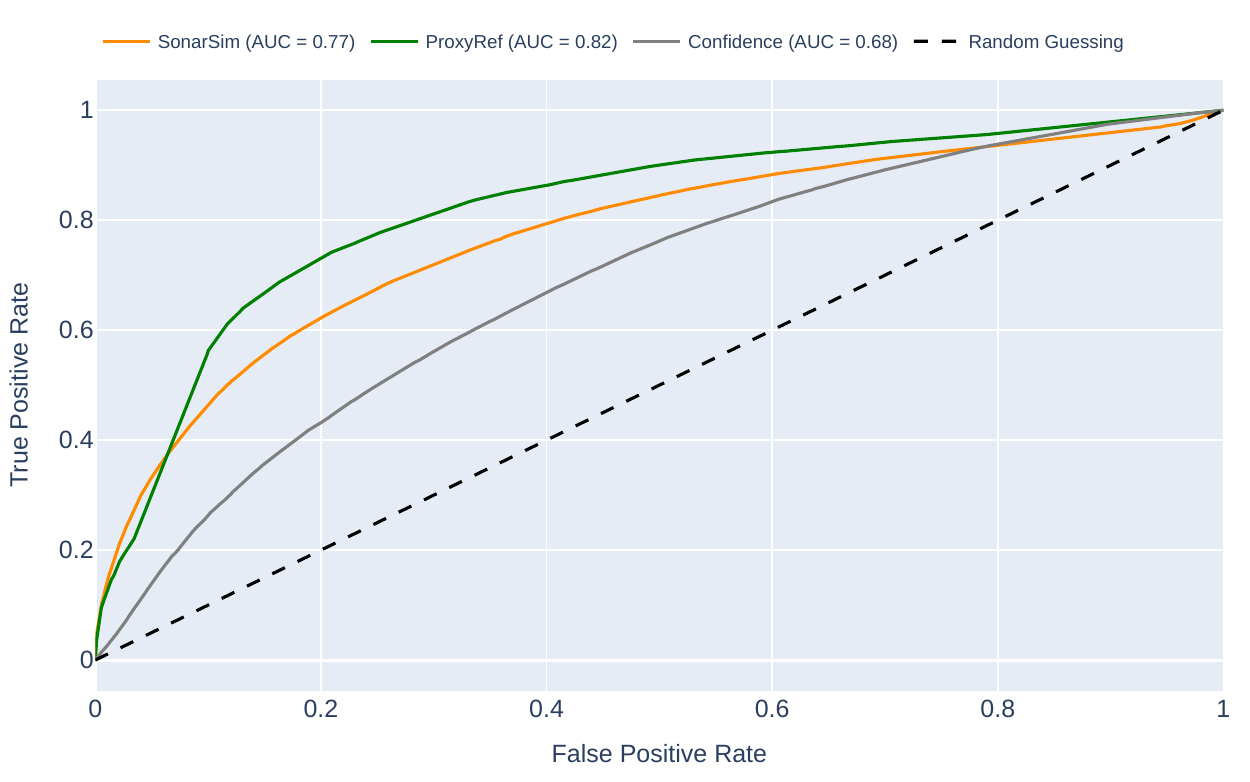}} % adjust width as needed
    \hspace{0.02\textwidth} % adjust the horizontal space between figures
    \subfigure[WER > 40]{\includegraphics[width=0.30\textwidth]{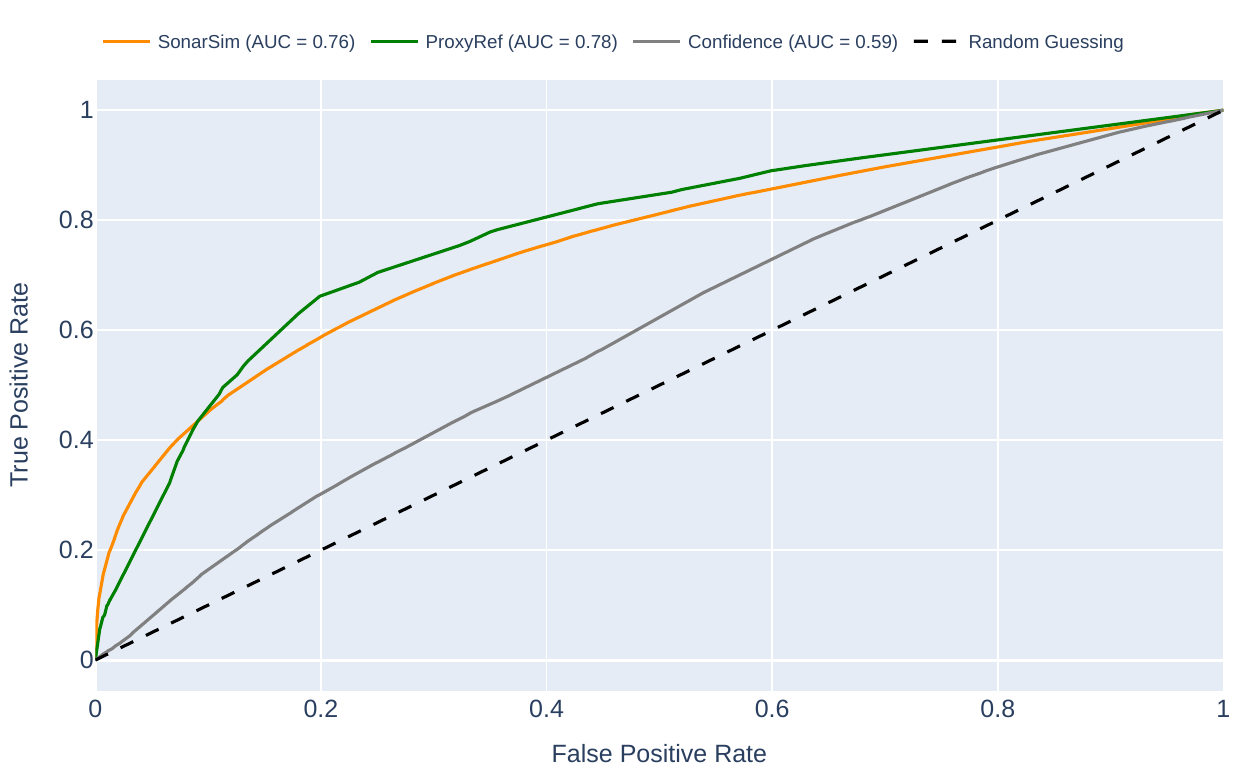}} % adjust width as needed
    \hspace{0.02\textwidth} % adjust the horizontal space between figures
    \subfigure[WER > 20]{\includegraphics[width=0.30\textwidth]{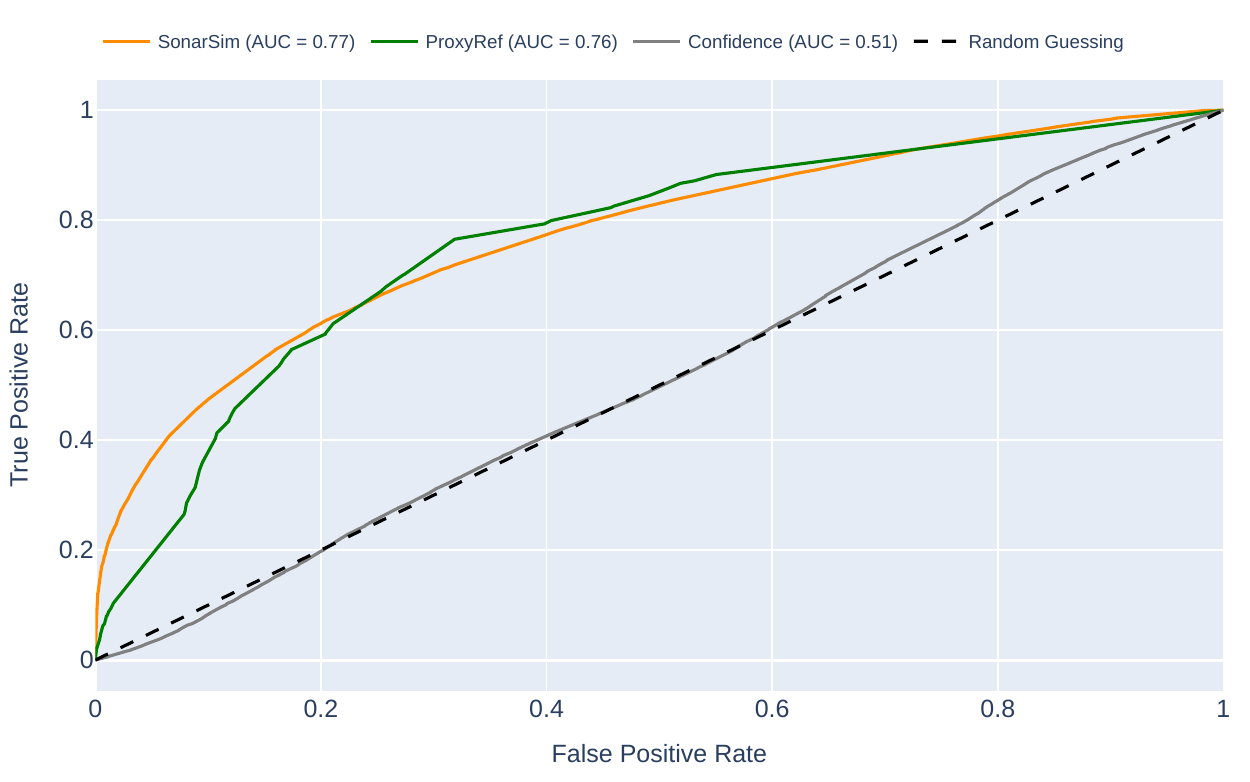}} % adjust width as needed
    \caption{Area under the curve (AUC) for detecting low-quality examples (WER > 20\%, 40\%, 80\%). The Y-axis represents the true positive rate (TPR), and the X-axis represents the false positive rate (FPR).}
    \label{fig:side-by-side}
\end{figure*}
We investigate the effectiveness of two of our best metrics for filtering low-quality pseudo-labels, specifically targeting instances with a WER higher than 80\%, 40\%, and 20\%. To assess their efficacy, we calculate the area under the curve (AUC) (as shown in Figure~\ref{fig:side-by-side}) for detecting low-quality examples. The results indicate that \textit{sonar-sim} achieves an AUC of 0.77 for detecting examples with a WER $> 80$, demonstrating reasonably high discriminative power in identifying low-quality labels. The \textit{proxy-ref} metric shows a slightly better performance, with an AUC of 0.82, indicating robust capability in distinguishing between high and low-quality pseudo-labels. In contrast, the confidence-based measure yielded an AUC of 0.68, which falls behind the other measures' discriminative power. These findings highlight sonar embeddings and the proxy reference-based measure as promising tools for improving the quality of pseudo-labels in scenarios where ground truth data is unavailable. 

\subsection{Experiments on Other Language.}

% \begin{table*}[ht!]
% \centering
% \resizebox{\textwidth}{!}{%
% \begin{tabular}{llccccc}
% \toprule
% \textbf{Evaluation} & \textbf{Dataset} & \textbf{W-L-v2} & \textbf{DW-16-16} & \textbf{DW-32-16} & \textbf{UDW-16-16-proxy} & \textbf{UDW-32-16-proxy} \\
% \midrule
% \multirow{3}{*}{\textbf{IID}} & \textbf{OpenBible}         & 101.3/44.4  & 59.14/\underline{14.0}  & \textbf{58.8/13.8}  & 59.2/14.0  & \underline{58.9}/14.1  \\
% & \textbf{CommonVoice17}     & 117.1/60.1  & 82.9/35.0  & \textbf{69.8/24.8}   & 75.6/29.2  & \underline{70.4/25.4}  \\
% & \textbf{ALFAA}             & 217.1/143.2 & 78.2/28.2  & \underline{74.4}/\textbf{25.7}   & 76.8/27.2   & \textbf{73.8}/\underline{26.5}  \\ 
% \midrule
% \multirow{3}{*}{\textbf{OOD}} & \textbf{DVoice}            & 214.6/144.6  & 124.4/74.1  & \textbf{110.2}/\underline{62.6} & \underline{110.7}/\textbf{62.4}  & 114.9/69.1  \\
% & \textbf{AMMI-LigAikuma}    & \textbf{46.7/13.0}   & 60.1/18.0   & \underline{51.8/14.4}  & 60.4/18.5  & 52.2/14.4  \\
% & \textbf{Fleurs}            & 54.6/\underline{14.8}   & 60.9/18.9  & \textbf{51.6/14.8}   & 58.9/18.5  & \underline{51.8}/14.9  \\
% \bottomrule
% \end{tabular}%
% }
% \caption{WER/CER comparison on Swahili. Best results are shown in \textbf{bold}. Second best results are \underline{underlined}. \KKcomment{move CER to apdx, make aligment more visually pleasing}}
% \label{tab:swahili_results}
% \end{table*}

%WER ONLY TABLE
\begin{table*}[ht!]
\centering
\resizebox{\textwidth}{!}{%
\begin{tabular}{llccccc}
\toprule
\multirow{2}{*}{Evaluation} & \multirow{2}{*}{Dataset} & \multicolumn{3}{c}{Baselines} & \multicolumn{2}{c}{Ours}\\
& & W-L-v2 & DW-16-16 & DW-32-16 & UDW-16-16$_{pr}$ & UDW-32-16$_{pr}$ \\
\midrule
\multirow{3}{*}{IID} & OpenBible         & 101.3  & 59.1  & \textbf{58.8}  & 59.2  & \underline{58.9}  \\
& CommonVoice17     & 117.1  & 82.9  & \textbf{69.8}   & 75.6  & \underline{70.4}  \\
& ALFAA             & 217.1 & 78.2  & \underline{74.4}   & 76.8   & \textbf{73.8}  \\ 
\midrule
\multirow{3}{*}{OOD} & DVoice            & 214.6  & 124.4  & \textbf{110.2} & \underline{110.7}  & 114.9  \\
& AMMI-LigAikuma    & \textbf{46.7}   & 60.1   & \underline{51.8}  & 60.4  & 52.2  \\
& Fleurs            & 54.6   & 60.9  & \textbf{51.6}   & 58.9  & \underline{51.8}  \\
\bottomrule
\end{tabular}%
}
\caption{WER (↓) results on the Swahili datasets. $pr$: using the proxy filtering method. Best results are shown in \textbf{bold}. Second best results are \underline{underlined}. WER scores are reported after normalization and removing diacritics.}
\label{tab:swahili_results}
\end{table*}
To further validate the effectiveness of our approach, we conduct experiments on Swahili, a low-resource language. We collect over 100 hours of labeled speech data from a variety of sources, namely OpenBible~\cite{meyer2022bibletts}, CommonVoice (Swahili subset)~\cite{ardila2020common}, ALFAA\footnote{\url{https://github.com/besacier/ALFFA_PUBLIC/tree/master/ASR/SWAHILI}}, DVoice~\cite{Gauthier2016CollectingRI}, AMMI-LIGAikuma\footnote{\url{https://github.com/besacier/AMMIcourse}}, and FLEURS (Swahili subset)~\cite{conneau2022fleurs}.

We distill two models, UDW-16-16 and UDW-32-16, using our best filtering method: \textit{proxy-ref}. The training data includes the train splits of OpenBible, CommonVoice, and ALFAA, and we evaluate the models on their respective test splits. We also test the models on three out-of-distribution (OOD) datasets: DVoice, AMMI-LigAikum, and FLEURS, which were not included in the training data.

% I\KKcomment{should we mention the amount of training vs test data?}.
% agree but might take some so possibly in rebuttal

We compare our distilled models to the teacher model and evaluate the performance of our unsupervised approach. The results show that our unsupervised distillation models perform on par with, or better than the supervised setup. Additionally, our distilled models outperform the teacher model by a significant margin on both familiar (IID) and novel (OOD) datasets, demonstrating the utility of our approach in extremely low-resource settings.
Specifically, the UDW-32-16 model achieves a WER/CER of 58.86/14.13\% on the IID OpenBible dataset, compared to the teacher model's 101.33/44.43\%. On the OOD dataset FLEURS, UDW-32-16 attains a WER/CER of 51.82/14.88, significantly outperforming the teacher model's 54.61/14.81. Across various datasets, our distilled models consistently outperform the teacher, with UDW-32-16 showing the best results overall.
Table~\ref{tab:swahili_results} presents the WER and CER scores for the different models and datasets.

These findings highlight the strength of our unsupervised data filtering approach, particularly in low-resource scenarios, where labeled data is scarce but the distilled models still perform robustly.

\section{Conclusion}\label{sec:conclusion}

% \AW{Taken from ACL work. Re-write it at the end.}
In this study, we explore methods for distilling large Whisper models into smaller, more efficient ones without relying on labeled data. Our filtering techniques bridge a gap in prior research and facilitate the creation of compact and effective speech recognition models for limited label settings. We show through a comprehensive evaluation that our models outperform both their teacher model and those using supervised distillation.
Our evaluation spans a diverse range of Arabic varieties, demonstrating their generalization to linguistic diversity and their competitive performance with SOTA models twice their size. Applying our approach to Swahili datasets further validates its effectiveness for different languages.
Notably, our model-based filtering methods (proxy and sonar) demonstrate superior robustness across linguistic variations. Moving forward, we aim to explore model-free approaches to further enhance the efficacy of model distillation, while including extremely low-resource languages and domains.

\section{Limitations}\label{sec:limitations}
% \AW{Taken from ACL work. Align/rephrase/repurpose for it for current work.}

In this study, we distill small Whisper models from relatively large ones via pseudo-labeling and unsupervised data filtering. Our distilled models are computationally efficient and maintain a performance similar to or better than the base teacher model and models trained in a supervised data filtering setup. Unlike~\citet{waheed2024distillar, gandhi2023distilwhisper}, our approach does not utilize any labeled data in the distillation process, making it directly applicable in data-scarce settings. However, despite these advantages, we acknowledge several limitations in our work, which we outline below.

\noindent\textbf{Efficiency.}
Our distilled models achieve 25-50\% compute efficiency relative to their larger counterparts while maintaining comparable performance. However, the training of these models requires significant computational resources.

Our main approach relies heavily on a robust reference model to serve as a proxy for filtering lower-quality pseudo labels. Specifically, we utilize SeamlessM4T-large-v2, a state-of-the-art model with 2.3 billion parameters, to generate proxy references which is then used to filter out low-quality data points. For similarity-based measures, we use SONAR~\cite{duquenne2023sonar} to generate multimodal embeddings from speech and pseudo labels. These embeddings provide contextual similarity which is then utilized to discard low-quality pseudo labels. We use AceGPT (7B), to compute the log-likelihood of the pseudo labels which is leveraged to filter out low-quality examples.

Although these measures allow attaining a performance on par or better than the supervised setup, it's important to highlight that each of these methodologies entails additional computational overhead.

\noindent\textbf{Multilinguality.} We use SeamlessM4T-large-v2 for generating proxy references, SONAR for generating multimodal embeddings, AceGPT (7B) for computing log-likelihood, and XTTS-v2 for generating synthetic speech. The multilingual capabilities of these models are crucial for effectively applying our techniques to a wide range of languages and dialects. However, a significant limitation of our approach is that it is constrained to languages supported by these models. This dependency restricts our ability to extend our distillation process to languages beyond the scope of the models' multilingual capacities.

\noindent\textbf{Evaluation.} Arabic is a linguistically rich and complex language with over 400 million speakers~\cite{400m,abdul2024nadi}, resulting in its wide range of varieties and dialects. We evaluate all the models on eleven different datasets representing different varieties, including five novel dialects collected and curated by native speakers and never seen before by any models. However, our varieties do not cover all Arabic-speaking regions. We aim to address this in future work by covering more varieties and dialects.

\noindent\textbf{Distillation Training Data.}
We distilled four variants of student models using 100K and 500K segments of which approximately 25\% are filtered. We see improvement going from 100K ($\approx$$100$ hours) to 500K ($\approx$$500$ hours) segments. As \citet{gandhi2023distilwhisper} shows, going over 1,000 hours results in a better model, we aim to study how distillation can be done under a low resource setting which is why we do not scale the data. Additionally, we also keep the WER threshold high (80) so that we remain close to a setting where no labeled data is available (even for filtering). It would be interesting, however, to see how distilled models may perform on unfiltered data in low-resource setting.

\noindent\textbf{Nature of Speech Data.}
Despite putting together a never-seen dataset of under-represented Arabic dialects, we realize that sourcing our data from television series renders its nature distant from speech spoken \textit{in the wild}. This type of content tends to be more ``theatrical'' and involves different elements such as background music and laughing tracks that do not accurately reflect regular conversational Arabic. Consequently, this could fail to accurately portray the performance of these models on real speech.

\section*{Acknowledgments}\label{sec:acknow}
We acknowledge support from Canada Research Chairs (CRC), the Natural Sciences and Engineering Research Council of Canada (NSERC; RGPIN-2018-04267), the Social Sciences and Humanities Research Council of Canada (SSHRC; 895-2020-1004; 895-2021-1008), Canadian Foundation for Innovation (CFI; 37771), Digital Research Alliance of Canada,\footnote{\href{https://alliancecan.ca}{https://alliancecan.ca}} and UBC Advanced Research Computing-Sockeye.\footnote{\href{https://arc.ubc.ca/ubc-arc-sockeye}{https://arc.ubc.ca/ubc-arc-sockeye}}
% \input{sections/9_ethics}

% Bibliography entries for the entire Anthology, followed by custom entries
%\bibliography{anthology,custom}
% Custom bibliography entries only
\bibliography{custom}

\begin{thebibliography}{50}
\providecommand{\natexlab}[1]{#1}

\bibitem[{Abdul-Mageed et~al.(2021)Abdul-Mageed, Elmadany, and Nagoudi}]{400m}
Muhammad Abdul-Mageed, AbdelRahim Elmadany, and El~Moatez~Billah Nagoudi. 2021.
\newblock \href {https://doi.org/10.18653/v1/2021.acl-long.551} {{ARBERT} {\&} {MARBERT}: Deep bidirectional transformers for {A}rabic}.
\newblock In \emph{Proceedings of the 59th Annual Meeting of the Association for Computational Linguistics and the 11th International Joint Conference on Natural Language Processing (Volume 1: Long Papers)}, pages 7088--7105, Online. Association for Computational Linguistics.

\bibitem[{Abdul-Mageed et~al.(2024)Abdul-Mageed, Keleg, Elmadany, Zhang, Hamed, Magdy, Bouamor, and Habash}]{abdul2024nadi}
Muhammad Abdul-Mageed, Amr Keleg, AbdelRahim Elmadany, Chiyu Zhang, Injy Hamed, Walid Magdy, Houda Bouamor, and Nizar Habash. 2024.
\newblock \href {https://doi.org/10.18653/v1/2024.arabicnlp-1.79} {{NADI} 2024: The fifth nuanced {A}rabic dialect identification shared task}.
\newblock In \emph{Proceedings of The Second Arabic Natural Language Processing Conference}, pages 709--728, Bangkok, Thailand. Association for Computational Linguistics.

\bibitem[{Abdul-Mageed et~al.(2020)Abdul-Mageed, Zhang, Elmadany, and Ungar}]{abdul-mageed-etal-2020-toward}
Muhammad Abdul-Mageed, Chiyu Zhang, AbdelRahim Elmadany, and Lyle Ungar. 2020.
\newblock \href {https://doi.org/10.18653/v1/2020.emnlp-main.472} {Toward micro-dialect identification in diaglossic and code-switched environments}.
\newblock In \emph{Proceedings of the 2020 Conference on Empirical Methods in Natural Language Processing (EMNLP)}, pages 5855--5876, Online. Association for Computational Linguistics.

\bibitem[{Al-Fetyani et~al.(2023)Al-Fetyani, Al-Barham, Abandah, Alsharkawi, and Dawas}]{e1qb-jv46-21}
Mohammad Al-Fetyani, Muhammad Al-Barham, Gheith Abandah, Adham Alsharkawi, and Maha Dawas. 2023.
\newblock \href {https://doi.org/10.1109/SLT54892.2023.10022652} {Masc: Massive arabic speech corpus}.
\newblock In \emph{2022 IEEE Spoken Language Technology Workshop (SLT)}, pages 1006--1013.

\bibitem[{Alharbi et~al.(2024)Alharbi, Alowisheq, Tüske, Darwish, Alrajeh, Alrowithi, Tamran, Ibrahim, Aloraini, Alnajim, Alkahtani, Almuasaad, Alrasheed, Alsubaie, and Alonaizan}]{10446243}
Sadeen Alharbi, Areeb Alowisheq, Zoltán Tüske, Kareem Darwish, Abdullah Alrajeh, Abdulmajeed Alrowithi, Aljawharah~Bin Tamran, Asma Ibrahim, Raghad Aloraini, Raneem Alnajim, Ranya Alkahtani, Renad Almuasaad, Sara Alrasheed, Shaykhah Alsubaie, and Yaser Alonaizan. 2024.
\newblock \href {https://doi.org/10.1109/ICASSP48485.2024.10446243} {Sada: Saudi audio dataset for arabic}.
\newblock In \emph{ICASSP 2024 - 2024 IEEE International Conference on Acoustics, Speech and Signal Processing (ICASSP)}, pages 10286--10290.

\bibitem[{Ali et~al.(2016)Ali, Bell, Glass, Messaoui, Mubarak, Renals, and Zhang}]{ali2019mgb2}
Ahmed Ali, Peter Bell, James Glass, Yacine Messaoui, Hamdy Mubarak, Steve Renals, and Yifan Zhang. 2016.
\newblock \href {https://doi.org/10.1109/SLT.2016.7846277} {The mgb-2 challenge: Arabic multi-dialect broadcast media recognition}.
\newblock In \emph{2016 IEEE Spoken Language Technology Workshop (SLT)}, pages 279--284.

\bibitem[{Ali et~al.(2019)Ali, Shon, Samih, Mubarak, Abdelali, Glass, Renals, and Choukri}]{9003960}
Ahmed Ali, Suwon Shon, Younes Samih, Hamdy Mubarak, Ahmed Abdelali, James Glass, Steve Renals, and Khalid Choukri. 2019.
\newblock \href {https://doi.org/10.1109/ASRU46091.2019.9003960} {The mgb-5 challenge: Recognition and dialect identification of dialectal arabic speech}.
\newblock In \emph{2019 IEEE Automatic Speech Recognition and Understanding Workshop (ASRU)}, pages 1026--1033.

\bibitem[{Ali et~al.(2017)Ali, Vogel, and Renals}]{ali2017speech}
Ahmed Ali, Stephan Vogel, and Steve Renals. 2017.
\newblock \href {https://doi.org/10.1109/ASRU.2017.8268952} {Speech recognition challenge in the wild: Arabic mgb-3}.
\newblock In \emph{2017 IEEE Automatic Speech Recognition and Understanding Workshop (ASRU)}, pages 316--322.

\bibitem[{Ardila et~al.(2020)Ardila, Branson, Davis, Kohler, Meyer, Henretty, Morais, Saunders, Tyers, and Weber}]{ardila2020common}
Rosana Ardila, Megan Branson, Kelly Davis, Michael Kohler, Josh Meyer, Michael Henretty, Reuben Morais, Lindsay Saunders, Francis Tyers, and Gregor Weber. 2020.
\newblock \href {https://aclanthology.org/2020.lrec-1.520/} {Common voice: A massively-multilingual speech corpus}.
\newblock In \emph{Proceedings of the Twelfth Language Resources and Evaluation Conference}, pages 4218--4222, Marseille, France. European Language Resources Association.

\bibitem[{Babu et~al.(2022)Babu, Wang, Tjandra, Lakhotia, Xu, Goyal, Singh, {von Platen}, Saraf, Pino, Baevski, Conneau, and Auli}]{babu2021xlsr}
Arun Babu, Changhan Wang, Andros Tjandra, Kushal Lakhotia, Qiantong Xu, Naman Goyal, Kritika Singh, Patrick {von Platen}, Yatharth Saraf, Juan Pino, Alexei Baevski, Alexis Conneau, and Michael Auli. 2022.
\newblock \href {https://doi.org/10.21437/Interspeech.2022-143} {Xls-r: Self-supervised cross-lingual speech representation learning at scale}.
\newblock In \emph{Interspeech 2022}, pages 2278--2282.

\bibitem[{Chang et~al.(2022)Chang, Yang, and Lee}]{distilhubert}
Heng-Jui Chang, Shu-wen Yang, and Hung-yi Lee. 2022.
\newblock \href {https://doi.org/10.1109/ICASSP43922.2022.9747490} {Distilhubert: Speech representation learning by layer-wise distillation of hidden-unit bert}.
\newblock In \emph{ICASSP 2022 - 2022 IEEE International Conference on Acoustics, Speech and Signal Processing (ICASSP)}, pages 7087--7091.

\bibitem[{Communication et~al.(2023)Communication, Barrault, Chung, Meglioli, Dale, Dong, Duquenne, ElSahar, Gong, Heffernan, Hoffman, Klaiber, Li, Licht, Maillard, Rakotoarison, Sadagopan, Wenzek, Ye, Akula, Chen, Hachem, Ellis, Gonzalez, Haaheim, Hansanti, Howes, Huang, Hwang, Inaguma, Jain, Kalbassi, Kallet, Kulikov, Lam, Li, Ma, Mavlyutov, Peloquin, Ramadan, Ramakrishnan, Sun, Tran, Tran, Tufanov, Vogeti, Wood, Yang, Yu, Andrews, Balioglu, Costa-juss{\`a}, Çelebi, Elbayad, Gao, Guzm'an, Kao, Lee, Mourachko, Pino, Popuri, Ropers, Saleem, Schwenk, Tomasello, Wang, Wang, and Wang}]{sm4t}
Seamless Communication, Lo{\"i}c Barrault, Yu-An Chung, Mariano~Cora Meglioli, David Dale, Ning Dong, Paul-Ambroise Duquenne, Hady ElSahar, Hongyu Gong, Kevin Heffernan, John Hoffman, Christopher Klaiber, Peng Li, Daniel Licht, Jean Maillard, Alice Rakotoarison, Kaushik~Ram Sadagopan, Guillaume Wenzek, Ethan Ye, Bapi Akula, Peng-Jen Chen, Naji~El Hachem, Brian Ellis, Gabriel~Mejia Gonzalez, Justin Haaheim, Prangthip Hansanti, Russ Howes, Bernie Huang, Min-Jae Hwang, Hirofumi Inaguma, Somya Jain, Elahe Kalbassi, Amanda Kallet, Ilia Kulikov, Janice Lam, Shang-Wen Li, Xutai Ma, Ruslan Mavlyutov, Benjamin Peloquin, M.L. Ramadan, Abinesh Ramakrishnan, Anna Sun, Ke~M. Tran, Tuan Tran, Igor Tufanov, Vish Vogeti, Carleigh Wood, Yilin Yang, Bo~Yu, Pierre~Yves Andrews, Can Balioglu, Marta~Ruiz Costa-juss{\`a}, Onur Çelebi, Maha Elbayad, Cynthia Gao, Francisco Guzm'an, Justine~T. Kao, Ann Lee, Alexandre Mourachko, Juan~Miguel Pino, Sravya Popuri, Christophe Ropers, Safiyyah Saleem, Holger Schwenk, Paden Tomasello,
  Changhan Wang, Jeff Wang, and Skyler Wang. 2023.
\newblock \href {https://api.semanticscholar.org/CorpusID:261064881} {Seamlessm4t: Massively multilingual\&multimodal machine translation}.

\bibitem[{Conneau et~al.(2020)Conneau, Khandelwal, Goyal, Chaudhary, Wenzek, Guzm{\'a}n, Grave, Ott, Zettlemoyer, and Stoyanov}]{conneau2020unsupervised}
Alexis Conneau, Kartikay Khandelwal, Naman Goyal, Vishrav Chaudhary, Guillaume Wenzek, Francisco Guzm{\'a}n, Edouard Grave, Myle Ott, Luke Zettlemoyer, and Veselin Stoyanov. 2020.
\newblock \href {https://doi.org/10.18653/v1/2020.acl-main.747} {Unsupervised cross-lingual representation learning at scale}.
\newblock In \emph{Proceedings of the 58th Annual Meeting of the Association for Computational Linguistics}, pages 8440--8451, Online. Association for Computational Linguistics.

\bibitem[{Conneau et~al.(2023)Conneau, Ma, Khanuja, Zhang, Axelrod, Dalmia, Riesa, Rivera, and Bapna}]{conneau2022fleurs}
Alexis Conneau, Min Ma, Simran Khanuja, Yu~Zhang, Vera Axelrod, Siddharth Dalmia, Jason Riesa, Clara Rivera, and Ankur Bapna. 2023.
\newblock \href {https://doi.org/10.1109/SLT54892.2023.10023141} {Fleurs: Few-shot learning evaluation of universal representations of speech}.
\newblock In \emph{2022 IEEE Spoken Language Technology Workshop (SLT)}, pages 798--805.

\bibitem[{Duquenne et~al.(2023)Duquenne, Schwenk, and Sagot}]{duquenne2023sonar}
Paul-Ambroise Duquenne, Holger Schwenk, and Beno{\^\i}t Sagot. 2023.
\newblock \href {https://arxiv.org/abs/2308.11466} {Sonar: Sentence-level multimodal and language-agnostic representations}.

\bibitem[{Ferraz et~al.(2024)Ferraz, Zanon~Boito, Brun, and Nikoulina}]{ferraz2024multilingualdistilwhisperefficientdistillation}
Thomas~Palmeira Ferraz, Marcely Zanon~Boito, Caroline Brun, and Vassilina Nikoulina. 2024.
\newblock \href {https://doi.org/10.1109/ICASSP48485.2024.10447520} {Multilingual distilwhisper: Efficient distillation of multi-task speech models via language-specific experts}.
\newblock In \emph{ICASSP 2024 - 2024 IEEE International Conference on Acoustics, Speech and Signal Processing (ICASSP)}, pages 10716--10720.

\bibitem[{Frantar et~al.(2022)Frantar, Ashkboos, Hoefler, and Alistarh}]{frantar2022gptq}
Elias Frantar, Saleh Ashkboos, Torsten Hoefler, and Dan Alistarh. 2022.
\newblock Gptq: Accurate post-training quantization for generative pre-trained transformers.
\newblock \emph{arXiv preprint arXiv:2210.17323}.

\bibitem[{Gandhi et~al.(2023)Gandhi, von Platen, and Rush}]{gandhi2023distilwhisper}
Sanchit Gandhi, Patrick von Platen, and Alexander~M. Rush. 2023.
\newblock \href {https://arxiv.org/abs/2311.00430} {Distil-whisper: Robust knowledge distillation via large-scale pseudo labelling}.
\newblock \emph{Preprint}, arXiv:2311.00430.

\bibitem[{Gauthier et~al.(2016)Gauthier, Besacier, Voisin, Melese, and Elingui}]{Gauthier2016CollectingRI}
Elodie Gauthier, Laurent Besacier, Sylvie Voisin, Michael Melese, and Uriel~Pascal Elingui. 2016.
\newblock \href {https://api.semanticscholar.org/CorpusID:216033883} {Collecting resources in sub-saharan african languages for automatic speech recognition: a case study of wolof}.
\newblock In \emph{International Conference on Language Resources and Evaluation}.

\bibitem[{Gou et~al.(2021)Gou, Yu, Maybank, and Tao}]{Gou_2021}
Jianping Gou, Baosheng Yu, Stephen~J. Maybank, and Dacheng Tao. 2021.
\newblock \href {https://doi.org/10.1007/s11263-021-01453-z} {Knowledge distillation: A survey}.
\newblock \emph{International Journal of Computer Vision}, 129(6):1789–1819.

\bibitem[{Halabi(2016)}]{halabi2016modern}
Nawar Halabi. 2016.
\newblock \emph{Modern standard Arabic phonetics for speech synthesis}.
\newblock Ph.D. thesis, University of Southampton.

\bibitem[{Hentschel et~al.(2024)Hentschel, Nishikawa, Komatsu, and Fujita}]{hentschel2024decodingparalleleffectiveknowledge}
Michael Hentschel, Yuta Nishikawa, Tatsuya Komatsu, and Yusuke Fujita. 2024.
\newblock \href {https://arxiv.org/abs/2401.11700} {Keep decoding parallel with effective knowledge distillation from language models to end-to-end speech recognisers}.
\newblock \emph{Preprint}, arXiv:2401.11700.

\bibitem[{Hinton et~al.(2015)Hinton, Vinyals, and Dean}]{hinton2015distillingknowledgeneuralnetwork}
Geoffrey Hinton, Oriol Vinyals, and Jeff Dean. 2015.
\newblock \href {https://arxiv.org/abs/1503.02531} {Distilling the knowledge in a neural network}.
\newblock \emph{Preprint}, arXiv:1503.02531.

\bibitem[{Hsu et~al.(2024)Hsu, Huang, and yi~Lee}]{hsu2024metawhisperspeechbasedmetaiclasr}
Ming-Hao Hsu, Kuan~Po Huang, and Hung yi~Lee. 2024.
\newblock \href {https://arxiv.org/abs/2409.10429} {Meta-whisper: Speech-based meta-icl for asr on low-resource languages}.
\newblock \emph{Preprint}, arXiv:2409.10429.

\bibitem[{Hsu et~al.(2021)Hsu, Bolte, Tsai, Lakhotia, Salakhutdinov, and Mohamed}]{hsu2021hubert}
Wei-Ning Hsu, Benjamin Bolte, Yao-Hung~Hubert Tsai, Kushal Lakhotia, Ruslan Salakhutdinov, and Abdelrahman Mohamed. 2021.
\newblock \href {https://arxiv.org/abs/2106.07447} {Hubert: Self-supervised speech representation learning by masked prediction of hidden units}.
\newblock \emph{Preprint}, arXiv:2106.07447.

\bibitem[{Hu et~al.(2020)Hu, Xie, Hong, and Tian}]{hu2020creating}
Hengtong Hu, Lingxi Xie, Richang Hong, and Qi~Tian. 2020.
\newblock Creating something from nothing: Unsupervised knowledge distillation for cross-modal hashing.
\newblock In \emph{Proceedings of the IEEE/CVF conference on computer vision and pattern recognition}, pages 3123--3132.

\bibitem[{Huang et~al.(2024)Huang, Yu, Zhu, Sun, Cheng, Song, Chen, Alharthi, An, He, Liu, Zhang, Chen, Li, Wang, Zhang, Sun, Wan, Li, and Xu}]{huang2024acegpt}
Huang Huang, Fei Yu, Jianqing Zhu, Xuening Sun, Hao Cheng, Dingjie Song, Zhihong Chen, Abdulmohsen Alharthi, Bang An, Juncai He, Ziche Liu, Zhiyi Zhang, Junying Chen, Jianquan Li, Benyou Wang, Lian Zhang, Ruoyu Sun, Xiang Wan, Haizhou Li, and Jinchao Xu. 2024.
\newblock \href {https://arxiv.org/abs/2309.12053} {Acegpt, localizing large language models in arabic}.
\newblock \emph{Preprint}, arXiv:2309.12053.

\bibitem[{Kim and Rush(2016)}]{kim-rush-2016-sequence}
Yoon Kim and Alexander~M. Rush. 2016.
\newblock \href {https://doi.org/10.18653/v1/D16-1139} {Sequence-level knowledge distillation}.
\newblock In \emph{Proceedings of the 2016 Conference on Empirical Methods in Natural Language Processing}, pages 1317--1327, Austin, Texas. Association for Computational Linguistics.

\bibitem[{Leviathan et~al.(2023)Leviathan, Kalman, and Matias}]{leviathan2023fastinferencetransformersspeculative}
Yaniv Leviathan, Matan Kalman, and Yossi Matias. 2023.
\newblock \href {https://arxiv.org/abs/2211.17192} {Fast inference from transformers via speculative decoding}.
\newblock \emph{Preprint}, arXiv:2211.17192.

\bibitem[{Lopes et~al.(2017)Lopes, Fenu, and Starner}]{data-free-kd}
Raphael~Gontijo Lopes, Stefano Fenu, and Thad Starner. 2017.
\newblock \href {https://arxiv.org/abs/1710.07535} {Data-free knowledge distillation for deep neural networks}.
\newblock \emph{CoRR}, abs/1710.07535.

\bibitem[{Malard et~al.(2023)Malard, Zaiem, and Algayres}]{malard2023bigmodelhardaudios}
Hugo Malard, Salah Zaiem, and Robin Algayres. 2023.
\newblock \href {https://arxiv.org/abs/2309.12712} {Big model only for hard audios: Sample dependent whisper model selection for efficient inferences}.
\newblock \emph{Preprint}, arXiv:2309.12712.

\bibitem[{Manohar et~al.(2018)Manohar, Ghahremani, Povey, and Khudanpur}]{ts-learning}
Vimal Manohar, Pegah Ghahremani, Daniel Povey, and Sanjeev Khudanpur. 2018.
\newblock \href {https://doi.org/10.1109/SLT.2018.8639635} {A teacher-student learning approach for unsupervised domain adaptation of sequence-trained asr models}.
\newblock In \emph{2018 IEEE Spoken Language Technology Workshop (SLT)}, pages 250--257.

\bibitem[{Meyer et~al.(2022)Meyer, Adelani, Casanova, Öktem, Weber, Kabongo, Salesky, Orife, Leong, Ogayo, Emezue, Mukiibi, Osei, Agbolo, Akinode, Opoku, Olanrewaju, Alabi, and Muhammad}]{meyer2022bibletts}
Josh Meyer, David~Ifeoluwa Adelani, Edresson Casanova, Alp Öktem, Daniel Whitenack~Julian Weber, Salomon Kabongo, Elizabeth Salesky, Iroro Orife, Colin Leong, Perez Ogayo, Chris Emezue, Jonathan Mukiibi, Salomey Osei, Apelete Agbolo, Victor Akinode, Bernard Opoku, Samuel Olanrewaju, Jesujoba Alabi, and Shamsuddeen Muhammad. 2022.
\newblock \href {https://arxiv.org/abs/2207.03546} {Bibletts: a large, high-fidelity, multilingual, and uniquely african speech corpus}.
\newblock \emph{Preprint}, arXiv:2207.03546.

\bibitem[{Mubarak et~al.(2021)Mubarak, Hussein, Chowdhury, and Ali}]{mubarak2021qasr}
Hamdy Mubarak, Amir Hussein, Shammur~Absar Chowdhury, and Ahmed Ali. 2021.
\newblock \href {https://arxiv.org/abs/2106.13000} {Qasr: Qcri aljazeera speech resource -- a large scale annotated arabic speech corpus}.
\newblock \emph{Preprint}, arXiv:2106.13000.

\bibitem[{Nayem et~al.(2023)Nayem, Xue, Chang, and Shanbhogue}]{Nayem2023}
Khandokar~Md. Nayem, Ran Xue, Ching-Yun~(Frannie) Chang, and Akshaya Vishnu~Kudlu Shanbhogue. 2023.
\newblock \href {https://www.amazon.science/publications/knowledge-distillation-on-joint-task-end-to-end-speech-translation} {Knowledge distillation on joint task end-to-end speech translation}.
\newblock In \emph{Interspeech 2023}.

\bibitem[{Pratap et~al.(2023)Pratap, Tjandra, Shi, Tomasello, Babu, Kundu, Elkahky, Ni, Vyas, Fazel-Zarandi, Baevski, Adi, Zhang, Hsu, Conneau, and Auli}]{pratap2023scaling}
Vineel Pratap, Andros Tjandra, Bowen Shi, Paden Tomasello, Arun Babu, Sayani Kundu, Ali Elkahky, Zhaoheng Ni, Apoorv Vyas, Maryam Fazel-Zarandi, Alexei Baevski, Yossi Adi, Xiaohui Zhang, Wei-Ning Hsu, Alexis Conneau, and Michael Auli. 2023.
\newblock \href {https://arxiv.org/abs/2305.13516} {Scaling speech technology to 1,000+ languages}.
\newblock \emph{Preprint}, arXiv:2305.13516.

\bibitem[{Radford et~al.(2023)Radford, Kim, Xu, Brockman, McLeavey, and Sutskever}]{whisper}
Alec Radford, Jong~Wook Kim, Tao Xu, Greg Brockman, Christine McLeavey, and Ilya Sutskever. 2023.
\newblock Robust speech recognition via large-scale weak supervision.
\newblock In \emph{International Conference on Machine Learning}, pages 28492--28518. PMLR.

\bibitem[{Sanh et~al.(2020)Sanh, Debut, Chaumond, and Wolf}]{sanh2020distilbert}
Victor Sanh, Lysandre Debut, Julien Chaumond, and Thomas Wolf. 2020.
\newblock \href {https://arxiv.org/abs/1910.01108} {Distilbert, a distilled version of bert: smaller, faster, cheaper and lighter}.
\newblock \emph{Preprint}, arXiv:1910.01108.

\bibitem[{Segal-Feldman et~al.(2024)Segal-Feldman, Shamsian, Navon, Hetz, and Keshet}]{segalfeldman2024whispermedusasearmultihead}
Yael Segal-Feldman, Aviv Shamsian, Aviv Navon, Gill Hetz, and Joseph Keshet. 2024.
\newblock \href {https://arxiv.org/abs/2409.15869} {Whisper in medusa's ear: Multi-head efficient decoding for transformer-based asr}.
\newblock \emph{Preprint}, arXiv:2409.15869.

\bibitem[{Shao et~al.(2023)Shao, Wang, Liu, Gong, Wang, and Qian}]{shao2023whisperkdq}
Hang Shao, Wei Wang, Bei Liu, Xun Gong, Haoyu Wang, and Yanmin Qian. 2023.
\newblock \href {https://arxiv.org/abs/2305.10788} {Whisper-kdq: A lightweight whisper via guided knowledge distillation and quantization for efficient asr}.
\newblock \emph{Preprint}, arXiv:2305.10788.

\bibitem[{Sun et~al.(2019)Sun, Cheng, Gan, and Liu}]{sun2019patient}
Siqi Sun, Yu~Cheng, Zhe Gan, and Jingjing Liu. 2019.
\newblock \href {https://arxiv.org/abs/1908.09355} {Patient knowledge distillation for bert model compression}.
\newblock \emph{Preprint}, arXiv:1908.09355.

\bibitem[{SYSTRAN()}]{faster_whisper}
SYSTRAN.
\newblock \href {https://github.com/SYSTRAN/faster-whisper} {faster-whisper: Faster whisper transcription with ctranslate2}.

\bibitem[{Talafha et~al.(2024)Talafha, Kadaoui, Magdy, Habiboullah, Chafei, El-Shangiti, Zayed, Tourad, Alhamouri, Assi et~al.}]{talafha2024casablanca}
Bashar Talafha, Karima Kadaoui, Samar Magdy, Mariem Habiboullah, Chafei Chafei, Ahmed El-Shangiti, Hiba Zayed, Mohamedou Tourad, Rahaf Alhamouri, Rwaa Assi, et~al. 2024.
\newblock Casablanca: Data and models for multidialectal arabic speech recognition.
\newblock In \emph{Proceedings of the 2024 Conference on Empirical Methods in Natural Language Processing}, pages 21745--21758.

\bibitem[{Talafha et~al.(2023)Talafha, Waheed, and Abdul-Mageed}]{talafha2023n}
Bashar Talafha, Abdul Waheed, and Muhammad Abdul-Mageed. 2023.
\newblock \href {https://doi.org/10.21437/Interspeech.2023-1044} {N-shot benchmarking of whisper on diverse arabic speech recognition}.
\newblock In \emph{Interspeech 2023}, pages 5092--5096.

\bibitem[{Tian et~al.(2022)Tian, Deng, Li, Ye, Cheng, Li, and Yan}]{tian22_interspeech}
Sanli Tian, Keqi Deng, Zehan Li, Lingxuan Ye, Gaofeng Cheng, Ta~Li, and Yonghong Yan. 2022.
\newblock \href {https://doi.org/10.21437/Interspeech.2022-775} {Knowledge distillation for ctc-based speech recognition via consistent acoustic representation learning}.
\newblock In \emph{Interspeech 2022}, pages 2633--2637.

\bibitem[{Waheed et~al.(2024)Waheed, Kadaoui, and Abdul-Mageed}]{waheed2024distillar}
Abdul Waheed, Karima Kadaoui, and Muhammad Abdul-Mageed. 2024.
\newblock \href {https://doi.org/10.18653/v1/2024.acl-long.680} {To distill or not to distill? on the robustness of robust knowledge distillation}.
\newblock In \emph{Proceedings of the 62nd Annual Meeting of the Association for Computational Linguistics (Volume 1: Long Papers)}, pages 12603--12621, Bangkok, Thailand. Association for Computational Linguistics.

\bibitem[{Yang et~al.(2023)Yang, Li, Zhang, and Woodland}]{yang2023knowledge}
Xiaoyu Yang, Qiujia Li, Chao Zhang, and Philip~C. Woodland. 2023.
\newblock \href {https://arxiv.org/abs/2303.10917} {Knowledge distillation from multiple foundation models for end-to-end speech recognition}.
\newblock \emph{Preprint}, arXiv:2303.10917.

\bibitem[{Yeo et~al.(2024)Yeo, Kim, Watanabe, and Ro}]{yeo2024visualspeechrecognitionlanguages}
Jeong~Hun Yeo, Minsu Kim, Shinji Watanabe, and Yong~Man Ro. 2024.
\newblock \href {https://arxiv.org/abs/2309.08535} {Visual speech recognition for languages with limited labeled data using automatic labels from whisper}.
\newblock \emph{Preprint}, arXiv:2309.08535.

\bibitem[{Zhang et~al.(2021)Zhang, Zhang, Liu, Cheng, and Li}]{zhang-etal-2021-matching}
Bo~Zhang, Xiaoming Zhang, Yun Liu, Lei Cheng, and Zhoujun Li. 2021.
\newblock \href {https://doi.org/10.18653/v1/2021.acl-long.421} {Matching distributions between model and data: Cross-domain knowledge distillation for unsupervised domain adaptation}.
\newblock In \emph{Proceedings of the 59th Annual Meeting of the Association for Computational Linguistics and the 11th International Joint Conference on Natural Language Processing (Volume 1: Long Papers)}, pages 5423--5433, Online. Association for Computational Linguistics.

\bibitem[{Zhang et~al.(2023)Zhang, Han, Qin, Wang, Bapna, Chen, Chen, Li, Axelrod, Wang, Meng, Hu, Rosenberg, Prabhavalkar, Park, Haghani, Riesa, Perng, Soltau, Strohman, Ramabhadran, Sainath, Moreno, Chiu, Schalkwyk, Beaufays, and Wu}]{usm}
Yu~Zhang, Wei Han, James Qin, Yongqiang Wang, Ankur Bapna, Zhehuai Chen, Nanxin Chen, Bo~Li, Vera Axelrod, Gary Wang, Zhong Meng, Ke~Hu, Andrew Rosenberg, Rohit Prabhavalkar, Daniel~S. Park, Parisa Haghani, Jason Riesa, Ginger Perng, Hagen Soltau, Trevor Strohman, Bhuvana Ramabhadran, Tara Sainath, Pedro Moreno, Chung-Cheng Chiu, Johan Schalkwyk, Françoise Beaufays, and Yonghui Wu. 2023.
\newblock \href {https://arxiv.org/abs/2303.01037} {Google usm: Scaling automatic speech recognition beyond 100 languages}.
\newblock \emph{Preprint}, arXiv:2303.01037.

\end{thebibliography}

\newpage
\appendix

% % \section{Example Appendix}
% \label{sec:appendix}
\section{Appendix}
\section{Dataset}~\label{appendix:dataset}
\subsection{SADA Dataset}~\label{appendix:sada_dataset}
Table~\ref{tab:sada_stats} summarizes the statistics of the SADA dataset used in our experiments.
\begin{table}[!hb]
\centering
\begin{tabular}{lcc}
\toprule
\multicolumn{1}{l}{Dialect} & \multicolumn{1}{l}{Test(S/D)} & \multicolumn{1}{l}{Valid (S/D)} \\ \midrule
Najdi                       & 1703/2.0709                                  & 2249/3.3155                                  \\
MTOS                        & 1320/4.8044                                  & 1048/3.82                                    \\
Khaliji                     & 1150/1.1308                                  & 679/0.6317                                   \\
Hijazi                      & 809/1.1202                                   & 528/0.6423                                   \\
Unknown                     & 762/0.8325                                   & 489/0.4861                                   \\
NA                          & 167/0.1341                                   & 2/0.0004                                     \\
MSA                         & 157/0.5406                                   & 54/0.1682                                    \\
Egyptian                    & 96/0.0865                                    & 45/0.0524                                    \\
Shamali                     & 18/0.0243                                    & -                                            \\
Yemeni                      & 7/0.0052                                     & 23/0.0349          \\                         
Levantine                   & - & 19/0.0137        \\ \midrule     
Total                       & 6189/10.75                                   & 5136/9.17   \\ \bottomrule                                
\end{tabular}
\caption{\label{tab:sada_stats}
SADA stats. S is the number of segments and D is the duration (in hours). \textbf{MTOS} - More than one speaker.
}
\end{table}

\begin{table}[!hb]
\centering
\begin{tabular}{lcc}
\toprule
\multicolumn{1}{l}{Dialect} & \multicolumn{1}{l}{Test(S/D)} & \multicolumn{1}{l}{Valid (S/D)} \\ \midrule
Najdi                       & 1703/2.0709                                  & 2249/3.3155                                  \\
MTOS                        & 1320/4.8044                                  & 1048/3.82                                    \\
Khaliji                     & 1150/1.1308                                  & 679/0.6317                                   \\
Hijazi                      & 809/1.1202                                   & 528/0.6423                                   \\
Unknown                     & 762/0.8325                                   & 489/0.4861                                   \\
NA                          & 167/0.1341                                   & 2/0.0004                                     \\
MSA                         & 157/0.5406                                   & 54/0.1682                                    \\
Egyptian                    & 96/0.0865                                    & 45/0.0524                                    \\
Shamali                     & 18/0.0243                                    & -                                            \\
Yemeni                      & 7/0.0052                                     & 23/0.0349          \\                         
Levantine                   & - & 19/0.0137        \\ \midrule     
Total                       & 6189/10.75                                   & 5136/9.17   \\ \bottomrule                                
\end{tabular}
\caption{\label{tab:sada_stats}
SADA stats. S is the number of segments and D is the duration (in hours). \textbf{MTOS} - More than one speaker.
}
\end{table}
\section{Experiments}~\label{appendix:experiments}
\subsection{CER Results}
We report the character error rates (CER) across different settings and datasets in Table~\ref{tab:results_main_cer}.
\begin{table}[h!]
\centering
\renewcommand{\arraystretch}{0.9} % Adjust row height
\setlength{\tabcolsep}{4pt} % Adjust column spacing
\resizebox{1.\linewidth}{!}{% Resize table to fit within the text width
\begin{tabular}{lcccccc}
\toprule
Model & Split & NJD & MTOS & KHLJ & HJZ & UNK \\
\midrule
\textbf{Baslines}\\
\midrule
W-FT & Test & 77.5 & 51.8 & 85.4 & 61.5 & 112.2 \\
     & Valid & 52.6 & 41.1 & 100.3 & 89.7 & 107.6 \\
SM4T-v1 & Test & 30.9 & 46.0 & 32.2 & 29.0 & 39.4 \\
       & Valid & 28.1 & 44.0 & 31.5 & 30.9 & 35.2 \\
SM4T-v2 & Test & 31.1 & 53.1 & 30.4 & 32.0 & 45.1 \\
       & Valid & 30.7 & 53.7 & 35.3 & 30.3 & 34.4 \\
W-M & Test & 65.8 & 79.3 & 77.0 & 59.7 & 122.2 \\
    & Valid & 56.9 & 75.1 & 62.9 & 52.0 & 106.5 \\
W-L-v2 & Test & 39.9 & 57.4 & 54.4 & 39.6 & 80.7 \\
      & Valid & 41.4 & 55.4 & 44.9 & 43.6 & 67.1 \\
W-L-v3 & Test & 31.6 & 53.7 & 44.1 & 38.6 & 61.3 \\
      & Valid & 30.2 & 47.7 & 39.2 & 27.2 & 49.2 \\
\midrule
DW-16-16 & Test & 30.8 & 47.6 & 32.7 & 30.7 & 39.8 \\
        & Valid & 31.4 & 44.7 & 35.2 & 32.8 & 39.8 \\
DW-32-16 & Test & 35.8 & 60.1 & 38.7 & 34.1 & 44.5 \\
        & Valid & 34.8 & 54.1 & 37.6 & 38.2 & 40.1 \\
DW-16-16++ & Test & 30.9 & 50.7 & 31.5 & 31.0 & 46.8 \\
          & Valid & 29.8 & 43.8 & 31.8 & 33.0 & 41.0 \\
DW-32-16++ & Test & 28.3 & 43.1 & 29.4 & 28.6 & 41.3 \\
          & Valid & 27.3 & 38.3 & 34.5 & 28.1 & 43.0 \\

\noalign{\vskip 0.3ex} % extra space above the dashline
\hdashline
\noalign{\vskip 0.6ex} % extra space below the dashline

No-Filter & & & & & & \\
$\quad-$ DW-16-16 & Test & 34.8 & 59.7 & 41.4 & 42.3 & 63.0 \\
                 & Valid & 38.9 & 53.4 & 41.9 & 37.7 & 54.3 \\
$\quad-$ DW-32-16 & Test & 42.8 & 63.9 & 47.0 & 45.9 & 63.2 \\
                 & Valid & 35.2 & 54.9 & 43.3 & 36.5 & 49.6 \\
\midrule
\textbf{Ours}\\
\midrule
UDW-16-16 & & & & & & \\
$\quad-$ proxy & Test & 35.5 & 55.6 & 38.9 & 39.6 & 52.0 \\
              & Valid & 34.0 & 50.9 & 39.1 & 37.1 & 41.2 \\
$\quad-$ sonar & Test & 35.8 & 30.3 & 55.7 & 38.8 & 36.8 \\
              & Valid & 35.9 & 39.3 & 52.4 & 38.7 & 36.7 \\

\noalign{\vskip 0.3ex} % extra space above the dashline
\hdashline
\noalign{\vskip 0.6ex} % extra space below the dashline

UDW-32-16 & & & & & & \\
$\quad-$ proxy & Test & 31.1 & 54.0 & 32.1 & 30.8 & 46.0 \\
              & Valid & 29.3 & 44.3 & \underline{29.1} & 28.6 & 36.6 \\
$\quad-$ sonar & Test & \underline{25.4} & \textbf{23.6} & 45.9 & 29.9 & \textbf{25.5} \\
              & Valid & 26.0 & \underline{26.7} & 44.1 & 30.3 & \underline{29.5} \\
              
\noalign{\vskip 0.3ex} % extra space above the dashline
\hdashline
\noalign{\vskip 0.6ex} % extra space below the dashline

UDW-16-16++ & & & & & & \\
$\quad-$ proxy & Test & 29.7 & 48.8 & 33.8 & 29.6 & 42.8 \\
              & Valid & 27.8 & 42.0 & 34.3 & 32.2 & 41.7 \\
$\quad-$ sonar & Test & 28.4 & 43.3 & 30.8 & 27.5 & 37.0 \\
              & Valid & 27.5 & 40.3 & 32.4 & 30.7 & 35.8 \\
UDW-32-16++ & & & & & & \\

\noalign{\vskip 0.3ex} % extra space above the dashline
\hdashline
\noalign{\vskip 0.6ex} % extra space below the dashline

$\quad-$ proxy & Test & \textbf{25.3} & 41.3 & 31.0 & \textbf{24.6} & 38.2 \\
              & Valid & \textbf{25.3} & 37.4 & 30.1 & 25.6 & 37.4 \\
$\quad-$ sonar & Test & 26.3 & 40.8 & \textbf{28.3} & \underline{24.8} & 34.5 \\
              & Valid & \textbf{25.3} & 37.0 & 30.2 & 29.8 & 34.2 \\
\bottomrule

\end{tabular}
}
\caption{ \label{tab:sada_top_five_dialects_results-cer}
CER ($\downarrow$) results on top five dialects/categories in SADA data. Best results are shown in \textbf{bold}. Second best results are \underline{underlined}. The scores are reported after normalization and removing diacritics.
}
% All baseline distilled models (dw-) are trained with a filtering threshold of 80 if not specified. We report the score on the test split of each dataset.
% Abbreviations. \textbf{W} - Whisper, \textbf{FT} - Finetuned, \textbf{M} - Medium, \textbf{L} - Large, \textbf{S} - Small, \textbf{U} - Unsupervised, \textbf{D} - Distil.
% This part of table caption is redundant.
\end{table}
\begin{table}[h!]
\centering
\renewcommand{\arraystretch}{0.8} % Adjust row height
\setlength{\tabcolsep}{4pt} % Adjust column spacing
\resizebox{1.\linewidth}{!}{% Resize table to fit within the text width
\begin{tabular}{lccccc}
\toprule
Model & NJD & MTOS & KHLJ & HJZ & UNK \\
\midrule
\textbf{Baselines}\\
\midrule
W-FT & 77.1 & 63.4 & 139.4 & 119.1 & 140.3 \\
SM4T-v1 & 51.9 & 68.7 & 61.7 & 54.2 & 62.3 \\
SM4T-v2 & 52.2 & 75.8 & 65.1 & 51.1 & \underline{59.8} \\
W-M & 80.4 & 102.8 & 89.5 & 72.9 & 127.7 \\
W-L-v2 & 60.9 & 72.9 & 67.7 & 64.5 & 68.0 \\
W-L-v3 & 49.3 & 65.5 & 67.5 & \textbf{46.5} & 67.7 \\
\midrule
DW-16-16 & 59.4 & 70.6 & 66.2 & 61.1 & 69.9 \\
DW-32-16 & 58.3 & 69.7 & 67.5 & 62.7 & 68.3 \\
DW-16-16++ & 56.8 & 72.0 & 62.3 & 60.2 & 75.2 \\
DW-32-16++ & 50.3 & 61.8 & 62.3 & 53.7 & 66.4 \\
\noalign{\vskip 0.3ex} % extra space above the dashline
\hdashline
\noalign{\vskip 0.6ex} % extra space below the dashline

No-Filter & & & & & \\
$\quad-$ DW-16-16 & 64.8 & 80.3 & 71.4 & 65.5 & 77.0 \\
$\quad-$ DW-32-16 & 57.9 & 73.7 & 68.6 & 56.8 & 72.3 \\
\midrule
\textbf{Ours}\\
\midrule
UDW-16-16 & & & & & \\
$\quad-$ proxy & 59.3 & 70.7 & 66.5 & 61.4 & 68.7 \\
$\quad-$ sonar & 64.8 & 67.5 & 78.1 & 69.9 & 65.3 \\
\noalign{\vskip 0.3ex} % extra space above the dashline
\hdashline
\noalign{\vskip 0.6ex} % extra space below the dashline

UDW-32-16 & & & & & \\
$\quad-$ proxy & 51.0 & 65.9 & \underline{58.1} & 53.7 & 64.9 \\
$\quad-$ sonar & \textbf{49.2} & \textbf{51.6} & 62.5 & 58.9 & \textbf{52.6} \\
\noalign{\vskip 0.3ex} % extra space above the dashline
\hdashline
\noalign{\vskip 0.6ex} % extra space below the dashline

UDW-16-16++ & & & & & \\
$\quad-$ proxy & 52.7 & 66.6 & 62.0 & 55.7 & 67.3 \\
$\quad-$ sonar & 53.6 & 64.4 & 62.3 & 55.7 & 63.7 \\
\noalign{\vskip 0.3ex} % extra space above the dashline
\hdashline
\noalign{\vskip 0.6ex} % extra space below the dashline

UDW-32-16++ & & & & & \\

$\quad-$ proxy & 49.0 & 60.4 & \textbf{57.6} & \underline{50.7} & 60.7 \\
$\quad-$ sonar & \underline{49.3} & \underline{58.8} & \underline{58.1} & 53.6 & 61.1 \\
\bottomrule

\end{tabular}
}
\caption{ \label{tab:sada_top_five_dialects_results}
WER ($\downarrow$) results on top five dialects/categories on the validation set of the SADA data. Best results are shown in \textbf{bold}. Second best results are \underline{underlined}. WER scores are reported after normalization and removing diacritics.}
\end{table}
\begin{table*}[h]
\centering
\Large
\renewcommand{\arraystretch}{1.0} % Adjust row height
\resizebox{1.\linewidth}{!}{% Adjust the table size to fit the width of the text

\begin{tabular}{lcccccccccccccc}
% Please add the following required packages to your document preamble:
% \usepackage{multirow}
\toprule
 \multirow{2}{*}{Model} & \multirow{2}{*}{Size}  & \multirow{2}{*}{CV15.0} & \multirow{2}{*}{MGB2} & \multirow{2}{*}{MGB3} & \multirow{2}{*}{MGB5} & \multirow{2}{*}{Fleurs} & \multicolumn{5}{c}{In-house Data} & \multirow{2}{*}{SADA}                                     \\
                                      &          &                       &                       &                       &                         & & ALG          & JOR         & PAL         & UAE          & YEM          \\ \midrule
   \textbf{Baselines}\\
   \midrule
  Amazon & -/- & -/- & -/- & -/- & -/- & -/- & 70.2 & 25.6 & 29.0 & 40.8 & 43.5 & -/- \\

\noalign{\vskip 0.3ex} % extra space above the dashline
\hdashline
\noalign{\vskip 0.6ex} % extra space below the dashline

 XLS-R & 0.96 & 39.4 & 53.1 & 61.6 & 68.0 & 43.9 & 67.0 & 61.4 & 61.1 & 64.6 & 63.6 & 68.3 \\

 HuBERT & 0.31 & 18.9 & 17.3 & \textbf{9.5} & 45.5 & 10.9 & 44.3 & 23.3 & 27.9 & 36.7 & 38.8 & 34.5 \\

 W-FT & 1.5 & 21.9 & \underline{8.1} & 26.9 & 62.3 & 3.4 & 69.6 & 37.2 & 35.4 & 69.1 & 64.8 & 65.7 \\ \midrule

 MMS-all & 1.0 & 80.9 & 13.4 & 34.6 & 45.9 & 6.3 & 78.0 & 55.4 & 75.1 & 78.1 & 76.6 & 	38.0  \\

\noalign{\vskip 0.3ex} % extra space above the dashline
\hdashline
\noalign{\vskip 0.6ex} % extra space below the dashline

 SM4T-M & 1.2 & 5.7 & 9.0 & 21.7 & 46.6 & 3.6 & 39.7 & 15.9 & 20.1 & 24.7 & 29.5 & 39.3 \\

 SM4T-L-v1 & 2.3 & 7.3 & 10.5 & 22.6 & 52.1 &  5.1 & 47.8 & 18.8 & 23.1 &  27.4 & 32.5 & 37.8 \\

 SM4T-L-v2 & 2.3 & \textbf{3.5} & 8.7 & 18.6 & 53.7 & 4.0 & 52.0 & 14.6 & 17.2 & 23.3 & 30.7 & 41.8 \\

\noalign{\vskip 0.3ex} % extra space above the dashline
\hdashline
\noalign{\vskip 0.6ex} % extra space below the dashline

 W-S & 0.24 & 16.4 & 24.7 & 51.9 & 164.8 & 8.7 & 84.7 & 32.9 & 36.3 & 59.7 & 66.7 & 103.6 \\

 W-M & 0.77 & 13.2 & 18.5 & 39.5 & 88.3 & 5.1 & 69.9 & 21.1 & 24.7 & 52.6 & 52.0 &  74.1 \\

 W-L-v2 & 1.5 & 7.8 & 15.3 & 33.0 & 68.9 & 3.6 & 71.7 & 17.0 & 22.3 & 38.2 & 45.5 & 51.2 \\

 W-L-v3 & 1.5 & 5.2 & \textbf{7.6} & \underline{17.3} & 44.6 & \underline{3.2} & 65.4 & 16.3 & 22.7 & 32.7 & 38.9 & 45.6 \\ \midrule

% % commenting this -> uncomment it if it's required for comparison
% & DW-8-8 & 0.44 & 12.3 & 17.8 & 36.6 & 53.0 & 11.4 & 48.2 & 29.0 & 33.0 & 38.4 & 41.5 & 52.3 \\

 DW-16-16 & 0.80 & 7.2 & 10.8 & 25.1 & 43.3 & 6.6 & 38.5 & 18.2 & 23.3 & 27.7 & 31.6 & 38.9 \\

 DW-32-16 & 1.12 & 5.9 & 8.9 & 21.4 & \textbf{40.4} & 4.8 & \underline{33.4} & 14.7 & 19.5 & 22.8 & 28.1 & 47.3 \\

% % commenting this -> uncomment it if it's required for comparison
% & DW-16-32 & 1.22 & 7.3 & 10.7 & 26.3 & 47.5 & 6.0 & 44.0 & 18.0 & 25.4 & 29.0 & 36.8 & 35.6 \\ 
\noalign{\vskip 0.3ex} % extra space above the dashline
\hdashline
\noalign{\vskip 0.6ex} % extra space below the dashline

 DW-16-16++ & 0.80 & 6.2 & 10.2 & 24.8 & 42.6 & 5.2 & 39.0 & 17.2 & 21.6 & 26.8 & 31.5 & 40.6  \\
 DW-32-16++ & 1.12 & 5.5 & 8.8 & 20.3 & \underline{40.6} & \textbf{3.1} & \textbf{33.3} & \textbf{13.4} & 18.8 & \underline{21.1} & \textbf{26.8} & 35.8 \\

\noalign{\vskip 0.3ex} % extra space above the dashline
\hdashline
\noalign{\vskip 0.6ex} % extra space below the dashline

 No-filter \\
 $\quad-$ DW-16-16 & 0.80 & 7.6 & 11.2 & 29.7 & 59.1 & 6.0 & 51.6 & 20.2 & 27.3 & 34.0 & 38.8 & 49.6 \\ 
 $\quad-$ DW-32-16 & 1.12 & 7.3 & 10.4 & 30.8 & 58.8 & 4.9 & 63.2 & 20.0 & 24.9 & 35.6 & 50.9 & 53.6\\
\midrule
\textbf{Ours}\\
\midrule

% \rowcolor{lightgreen}
 UDW-16-16 & 0.80 \\
% \rowcolor{lightgreen}
 $\quad-$ nll &  & 8.15 & 11.26 & 27.98 & 55.25 & 6.26 & 41.4 & 25.7 & 20.52 & 35.96 & 49.2
 & 55.0 \\ 
% \rowcolor{lightgreen}
 $\quad-$ pesq &  & 8.41 & 12.11 & 27.69 & 54.88 & 6.89 & 40.34 & 27.41 & 20.16 & 32.55 & 44.17 & 50.1 \\ 
% \rowcolor{lightgreen}
 $\quad-$ entropy & & 8.1 & 12.17 & 31.24 & 56.64 & 6.4 & 48 & 22.81 & 27.67 & 37.85 & 52.56 & 61.8  \\

% \rowcolor{lightgreen}
 $\quad-$ conf & & 7.83 & 11.87 & 27.85 & 50.73 & 6.12 & 43.94 & 20.29 & 25.52 & 31.75 & 39.27 & 49.2  \\

% \rowcolor{lightgreen}
 $\quad-$ proxy & &7.48 & 11.39 & 26.36 & 49.97 & 7.5 & 42.15 & 23.69 & 19.66 & 30.93 & 41.94 & 46.2 \\ 
% \rowcolor{lightgreen}
 $\quad-$ sonar & & 8.04 & 11.86 & 28.66 & 49.21 & 7.06 & 43.48 & 22.61 & 27.43 & 32.89 & 36.13 & 45.6 \\ 

\noalign{\vskip 0.3ex} % extra space above the dashline
\hdashline
\noalign{\vskip 0.6ex} % extra space below the dashline

% \rowcolor{lightgreen}
 UDW-32-16 & 1.12 \\
% \rowcolor{lightgreen}
 $\quad-$ nll & & 6.24 & 10.12 & 25.39 & 55.53 & 4.47 & 35.85 & 20.88 & \underline{16.04} & 30.49 & 41.38
 & 46.1 \\ 
% \rowcolor{lightgreen}
 $\quad-$ pesq & & 7.5 & 10.6 & 26.4 & 51.0 & 5.3 & 43.2 & 17.2 & 22.7 & 30.9 & 36.4 & 48.1 \\

% \rowcolor{lightgreen}
 $\quad-$ entropy & & 6.53 & 10.34 & 28.71 & 66.87 & 4.34 & 84.02 & 21.07 & 31.08 & 44.23 & 54.25
 & 52.3  \\

% \rowcolor{lightgreen}
 $\quad-$ conf & & 6.46 & 9.79 & 23.42 & 48.61 & 4.73 & 36.08 & 21.99 & 25.82 & \textbf{17.7} & 42.82 & 41.3  \\

% \rowcolor{lightgreen}
 $\quad-$ proxy & & 6.17 & 9.87 & 22.45 & 46.05 & 6.23 & 36.11 & 15.6 & 20.88 & 25.69 & 29.49 & 41.9 \\ 
% \rowcolor{lightgreen}
 $\quad-$ sonar & & 5.62 & 8.98 & 23.4 & 46.97 & 4.41 & 37 & 15.11 & 19.56 & 24.29 & 28.24
 & 35.6 \\ 

\noalign{\vskip 0.3ex} % extra space above the dashline
\hdashline
\noalign{\vskip 0.6ex} % extra space below the dashline

% \rowcolor{lightgreen}

 UDW-16-16++ & 0.80 \\
% \rowcolor{lightgreen}
 $\quad-$ proxy & & \underline{4.8} & 10.4 & 24.3 & 48.0 & 4.8 & 41.6 & 16.3 & 21.1 & 27.5 & 33.0 & 35.3 \\ 
% \rowcolor{lightgreen}
 $\quad-$ sonar & & 6.1 & 9.8 & 24.2 & 47.1 & 5.0 & 38.9 & 17.2 & 21.4 & 26.2 & 29.5 & 33.96 \\
% \rowcolor{lightgreen}
 UDW-32-16-++ & 1.12 \\
% \rowcolor{lightgreen}
 $\quad-$ proxy & & 5.8 & 9.2 & 21.1 & 44.1 & 4.2 & 37.1 & \underline{14.5} & 20.1 & 23.1 & 29.2 & \textbf{31.4}  \\ 
% \rowcolor{lightgreen}
 $\quad-$ sonar & & 5.3 & 9.9 & 22.4 & 44.2 & 3.9 & 34.6 & 19.2 & \textbf{15.1} & 23.2 & \underline{27.0} & \underline{33.4} \\

\bottomrule 
\end{tabular}
}
\caption{ \label{tab:results_main_cer}
CER ($\downarrow$) scores after normalization and removing diacritics. All baseline distilled models (DW-) are trained with a filtering threshold of 80 if not specified. Best results are shown in \textbf{bold}. Second best results are \underline{underlined}. We report the score on the test split of each dataset.
Abbreviations. \textbf{W} - Whisper, \textbf{FT} - Finetuned, \textbf{M} - Medium, \textbf{L} - Large, \textbf{S} - Small, \textbf{U} - Unsupervised, \textbf{D} - Distil, \textbf{nll} - negative log likelihood, \textbf{conf} - confidence score.
}
\end{table*}

\begin{table}[h!]
\centering
\setlength{\tabcolsep}{4pt}
\small 
\renewcommand{\arraystretch}{1.1}   %to squeeze table content low value means squeeze more
\resizebox{1.\linewidth}{!}{% reduce the text size, lower means smaller text size

\begin{tabular}{lccccc}
\toprule
\multirow{2}{*}{Model} & \multicolumn{2}{c}{Bench} & \multicolumn{2}{c}{SADA2022} & \multirow{2}{*}{IH} \\
                       & Test          & Valid         & Test         & Valid         &     \\ \midrule

\textbf{Baselines}\\
\midrule
HuBERT & 20.4 & 22.7 & 34.5 & 31.9 & 34.2 \\ 

W-FT & 24.5 & 28.6 & 65.7 & 56.2 & 55.2 \\
\noalign{\vskip 0.3ex} % extra space above the dashline
\hdashline
\noalign{\vskip 0.6ex} % extra space below the dashline
           
SM4T-v1 & 19.5 & 21.2 & 37.8  & 35.6 & 29.9 \\

SM4T-v2 & 17.7 & 19.4  & 41.8 & 40.8 & 27.6 \\

W-M & 32.9 & 38.0 & 74.1 & 66.7  & 44.1 \\

W-L-v2 & 25.7 & 29.7 & 51.2 & 48.5  & 38.9  \\

W-L-v3 & \textbf{15.6} & \textbf{17.1} & 45.6 & 39.3 & 35.2 \\

\midrule

DW-16-16 & 18.6 & 20.5  & 38.9 & 37.8 & 27.8 \\

DW-32-16 & 16.3 & \underline{17.9}  & 47.3  & 43.9 & \underline{23.7} \\

DW-16-16++ & 17.8 & 19.4 & 40.6 & 36.6 & 27.2 \\
DW-32-16++ & \underline{15.7} & \textbf{17.1} & 35.8 & 33.4 & \textbf{22.7} \\ 

\noalign{\vskip 0.3ex} % extra space above the dashline
\hdashline
\noalign{\vskip 0.6ex} % extra space below the dashline

No-Filter \\
$\quad-$ DW-16-16 & 22.7 & 25.1 & 49.6 & 45.9 & 34.4  \\ 
$\quad-$ DW-32-16 & 22.5 & 25.8 & 53.6 & 45.0 & 38.9  \\ 

\midrule      
\textbf{Ours}\\
\midrule
UDW-16-16 \\
% $\quad-$ entropy & 0.0 & 0.0 & 0.0 & 0.0 & 0.0  \\ 
% $\quad-$ nll & 0.0 & 0.0 & 0.0 & 0.0 & 0.0 \\ 
% $\quad-$ pesq & 0.0 & 0.0 & 0.0 & 0.0 & 0.0  \\ 
$\quad-$ proxy & 20.5 & 22.1 & 46.2 & 42.2 & 31.7  \\ 
$\quad-$ sonar & 21.0 & 22.8 & 45.6 & 43.5 & 32.5  \\

\noalign{\vskip 0.3ex} % extra space above the dashline
\hdashline
\noalign{\vskip 0.6ex} % extra space below the dashline

UDW-32-16 \\
% $\quad-$ entropy & 0.0 & 0.0 & 0.0 & 0.0 & 0.0   \\ 
% $\quad-$ nll & 0.0 & 0.0 & 0.0 & 0.0 & 0.0  \\ 
% $\quad-$ pesq & 0.0 & 0.0 & 0.0 & 0.0 & 0.0 \\ 
$\quad-$ proxy & 18.2 & 19.6 & 41.9 & 36.1 & 25.6  \\ 
$\quad-$ sonar & 17.9 & 19.7 & 35.6 & 34.7 & 24.8  \\

\noalign{\vskip 0.3ex} % extra space above the dashline
\hdashline
\noalign{\vskip 0.6ex} % extra space below the dashline

UDW-16-16++ \\

$\quad-$ proxy & 18.4 & 21.2 & 39.2 & 35.3 & 27.9  \\ 
$\quad-$ sonar & 18.4 & 19.9 & 35.4 & 34.0 & 26.6  \\

\noalign{\vskip 0.3ex} % extra space above the dashline
\hdashline
\noalign{\vskip 0.6ex} % extra space below the dashline

UDW-32-16++ \\
$\quad-$ proxy & 17.1 & 18.7 & \underline{33.6} & \underline{31.4} & 23.8  \\ 
$\quad-$ sonar & 16.9 & 18.5 & \textbf{33.1} & \textbf{31.3} & 24.8  \\

\bottomrule
\end{tabular}
}
\caption{ \label{tab:results_average-cer}
Average CER ($\downarrow$) across different evaluation datasets. Bench: CV15.0, FLEURS and the three MGBs. Best results are shown in \textbf{bold}. Second best results are \underline{underlined}. The scores are reported after normalization and removing diacritics.
}
% All baseline distilled models (dw-) are trained with a filtering threshold of 80 if not specified. We report the score on the test split of each dataset.
% Abbreviations. \textbf{W} - Whisper, \textbf{FT} - Finetuned, \textbf{M} - Medium, \textbf{L} - Large, \textbf{S} - Small, \textbf{U} - Unsupervised, \textbf{D} - Distil.
% This part of table caption is redundant.
\end{table}
\begin{table*}[ht!]
\centering
\resizebox{\textwidth}{!}{%
\begin{tabular}{llccccc}
\toprule
\multirow{2}{*}{Evaluation} & \multirow{2}{*}{Dataset} & \multicolumn{3}{c}{Baselines} & \multicolumn{2}{c}{Ours}\\
& & W-L-v2 & DW-16-16 & DW-32-16 & UDW-16-16$_{pr}$ & UDW-32-16$_{pr}$ \\
\midrule
\multirow{3}{*}{IID} & OpenBible         & 44.4  & \underline{14.0}  & \textbf{13.8}  & 14.0  & 14.1  \\
& CommonVoice17     & 60.1  & 35.0  & \textbf{24.8}   & 29.2  & \underline{25.4}  \\
& ALFAA             & 143.2 & 28.2  & \textbf{25.7}   & 27.2   & \underline{26.5}  \\ 
\midrule
\multirow{3}{*}{OOD} & DVoice            & 144.6  & 74.1  & \underline{62.6} & \textbf{62.4}  & 69.1  \\
& AMMI-LigAikuma    & \textbf{13.0}   & 18.0   & \underline{14.4}  & 18.5  & 14.4  \\
& Fleurs            & \underline{14.8}   & 18.9  & \textbf{14.8}   & 18.5  & 14.9  \\
\bottomrule
\end{tabular}%
}
\caption{CER (↓) results on the Swahili datasets. $pr$: using the proxy filtering method. Best results are shown in \textbf{bold}. Second best results are \underline{underlined}. WER scores are reported after normalization and removing diacritics}
\label{tab:swahili_results-cer}
\end{table*}
\begin{table*}[]
    \centering

    \begin{tabular}{lccccccccc}
    \toprule
    Model & ALG & EGY & JOR & MAU & MOR & PAL & UAE & YEM & AVG\\
    \midrule
    \textbf{Baselines}\\
    \midrule
    SM4T-v2 & 53.48 & \textbf{26.12} & \textbf{13.15} & 52.20 & 54.96 & \textbf{18.20} & \textbf{22.71} & \textbf{27.07} & 34.44\\
    W-L-v2 & 58.63 & 30.28 & 20.37 & 79.66  & 63.21  & 25.70 & 38.06 & 51.49 & 46.83\\
    \midrule
    DW-16-16 & \underline{40.08} & 31.80 & 19.11 & \underline{49.83} & \underline{42.16} & 24.10 & 26.99 & 30.53 & 33.64\\
    DW-32-16 & 44.45 & 32.80 & 19.27 & 49.95 & 43.46 & 26.43 & 26.26 & 34.03 & 35.12\\
    No-Filter & & & & & & & & & \\
    $\quad-$ DW-32-16 & 61.50 & 43.52 & 18.41 & 64.19 & 51.36 & 29.44 & 36.97 & 41.75 & 43.95\\
    \midrule
    \textbf{Ours}\\
    \midrule
    UDW-16-16 & & & & & & & & & \\
    $\quad-$ proxy & 48.30 & 39.79 & 20.21 & 53.06 & 45.92 & 25.69 & 29.15 & 37.13 & 38.01\\
    $\quad-$ sonar & 43.94 & 36.24 & 23.60 & 55.10 & 50.14 & 28.77 & 31.05 & 34.65 & 38.63\\
    \hdashline
    UDW-32-16 & & & & & & & & & \\
    $\quad-$ proxy & 40.72 & 29.89 & 16.23 & \textbf{47.03} & \textbf{41.45} & 23.42 & 23.72 & 27.26 & \underline{31.80}\\
    $\quad-$ sonar & \textbf{38.34} & \underline{28.61} & \underline{16.02} & 50.02 & 44.94 & \underline{19.87} & \underline{23.13} & \underline{27.17} & \textbf{31.79}\\
    \bottomrule
    \end{tabular}%
    \caption{CER results on the Casablanca dataset. The best results are shown in bold. The second-best results are underlined. CER ($\downarrow$) scores are reported after normalization and removing diacritics.
We report the score on the test split of each dataset.}
    \label{tab:casablanca_results_cer}
\end{table*}

\subsection{Training Parameters}~\label{appendix:hyperparameters}
Table~\ref{training-parameters} lists the hyperparameters used for training our models across all experiments.
\begin{table*}[]
\centering
\begin{tabular}{ll}
\toprule
Parameter                     & Value                  \\
\midrule
\texttt{warmup\_steps}                 & $50$                     \\
\texttt{learning\_rate}                & $0.0001$                 \\
\texttt{lr\_scheduler\_type}           & \texttt{constant\_with\_warmup} \\
\texttt{batch\_size}                   & $128$                    \\
\texttt{max\_label\_length}            & $225$                    \\
\texttt{gradient\_accumulation\_steps} & $1$                      \\
\texttt{dtype}                         & \texttt{bfloat16}      \\

\bottomrule
\end{tabular}
\caption{\label{training-parameters} Training parameters. All the training parameters are the default ones provided in Huggingface Seq2SeqTrainingArguments unless specified otherwise in this table.
}
\end{table*}
\subsection{Results}\label{appendix:results}
We present additional experimental results evaluating orthographic variants in Table~\ref{tab:ortho-results-appendix}.
% Please add the following required packages to your document preamble:
% \usepackage{multirow}
% for cdashline \cdashline{1-1}[1pt/2pt]{}
\newcolumntype{G}{>{\columncolor{lightgreen}}c}

\begin{table*}[h!]
\centering
\Large 
\renewcommand{\arraystretch}{1.1}   %to squeeze table content low value means squeeze more
\resizebox{1.\linewidth}{!}{% reduce the text size, lower means smaller text size
\begin{tabular}{lcccccccccccc}
% Please add the following required packages to your document preamble:
% \usepackage{multirow}
\toprule
 \multirow{2}{*}{Model} & \multirow{2}{*}{Size}  & \multirow{2}{*}{CV15.0} & \multirow{2}{*}{MGB2} & \multirow{2}{*}{MGB3} & \multirow{2}{*}{MGB5} & \multirow{2}{*}{Fleurs} & \multicolumn{5}{c}{In-house Data} & \multirow{2}{*}{SADA}                                     \\
                                      &          &                       &                       &                       &                         & & ALG          & JOR         & PAL         & UAE          & YEM          \\ \midrule
% \hline
% \hline
\textbf{Baslines}\\
\midrule
 Amazon & -/- & -/- & -/- & -/- & -/- & -/- & 88.0/71.6 & 59.2/29.1 & 63.4/32.2 & 71.1/44.3 & 77.4/47.7 & - \\ 
\hdashline

 XLS-R & 0.96 & 92.7/46.7 & 97.7/54.5 & 99.1/64.5 & 99.6/70.1 & 95.1/45.4 & 99.7/68.0 & 99.3/62.9 & 99.2/62.8 & 99.5/66.4 & 99.7/66.4 & 99.6/69.5 \\

 HuBERT &0.31 & 76.5/31.0 & 59.4/20.3 & \textbf{43.3/16.5} & 95.0/48.7 & 48.9/14.4 & 96.2/45.6 & 70.6/25.4 & 81.5/31.4 & 87.9/39.9 & 91.3/40.8 & 81.3/37.1 \\

 W-FT & 1.5 & 70.0/33.8 & 29.4/10.9 & 60.1/32.2 & 105.0/64.3 & 28.7/7.3 & 114.5/70.3 & 75.1/39.0 & 81.3/38.7 & 113.7/70.9 & 110.1/65.6 & 101.4/67.6 \\ \hdashline

 MMS-all & 1.0 & 106.0/82.5 & 40.3/14.0 & 77.7/38.1 & 90.4/48.5 & 28.8/7.8 & 100.2/77.8 & 91.5/56.2 & 100.0/75.8 & 100.1/78.4 & 100.1/76.8 & 79.8/39.1 \\ \hdashline

 SM4T-M & 1.2 & 42.3/18.2 & 28.1/11.2 & 50.2/26.8 & 88.2/50.8 & 19.5/6.0 & 84.5/42.8 & 55.2/18.7 & 63.0/23.0 & 68.0/28.1 & 79.4/34.5 & 73.2/42.8 \\

 SM4T-L-v1 & 2.3 & 44.2/19.1 & 25.9/11.7 & 52.5/27.6 & 92.8/55.9 & 22.6/7.6 & 89.7/50.3 & 59.1/21.7 & 64.7/25.8 & 69.0/30.3 & 81.5/37.0 & 72.4/40.8 \\

 SM4T-L-v2 & 2.3 & \textbf{37.7/15.8} & 22.4/9.9 & 46.7/23.9 & 92.1/58.4 & 19.8/6.5 & 94.8/55.2 & 51.3/17.6 & \textbf{58.5/20.1} & 65.6/26.9 & 80.6/35.5 & 72.2/44.4 \\ \hdashline

 W-S & 0.24 & 68.9/31.8 & 49.5/25.7 & 84.8/55.4 & 228.6/164.5 & 33.4/10.3 & 129.15/87.85 & 75.25/36.55 & 79.73/39.3 & 103.83/63 & 112.69/70.69 & 144.5/106.6 \\ 

 W-M & 0.77 & 55.1/24.2 & 37.6/19.6 & 71.5/43.7 & 129.7/89.4 & 24.0/7.1 & 103.9/71.4 & 59.0/23.9 & 66.8/27.6 & 90.7/55.7 & 95.2/56.2 & 106.0/76.3 \\

 W-L-v2 & 1.5 & 46.9/19.6 & 33.7/16.9 & 60.6/37.7 & 101.1/71.1 & 19.7/5.6 & 106.9/74.6 & 51.2/19.6 & 60.2/25.2 & 73.2/41.2 & 86.9/50.1 & 78.0/53.5 \\

 W-L-v3 & 1.5 & 43.2/16.9 & \textbf{20.4/8.6} & 44.6/22.5 & 82.0/47.7 & \textbf{16.4/4.8} & 103.8/68.9 & 52.7/18.9 & 64.3/26.4 & 72.3/35.9 & 86.0/43.3 & 74.6/47.9 \\ \hline

% % commenting this -> uncomment it if it's required for comparison
% & DW-8-8 & 0.44 & 55.0/23.2 & 44.4/19.2 & 69.2/40.4 & 91.0/55.5 & 36.1/13.3 & 91.5/49.6 & 71.4/31.2 & 78.4/35.6 & 82.5/41.2 & 87.5/44.9 & 83.8/53.9 \\

 DW-16-16 & 0.80 & 48.0/18.9 & 33.2/12.5 & 57.1/29.6 & 84.1/46.2 & 26.2/8.5 & 83.8/40.2 & 57.8/20.5 & 68.2/26.2 & 72.0/31.0 & 80.0/35.6 & 72.0/40.9 \\

 DW-32-16 & 1.12 & 45.6/17.7 & 27.7/10.3 & 51.2/26.1 & 80.9/\textbf{43.4} & 22.0/6.6 & \textbf{80.5/35.1} & 52.6/17.1 & 62.9/22.4 & 66.7/26.3 & 77.3/32.6 & 72.3/49.2 \\

% % commenting this -> uncomment it if it's required for comparison
% & DW-16-32 & 1.22 & 47.4/18.8 & 29.9/11.8 & 55.5/30.7 & 84.7/50.2 & 26.0/7.9 & 84.8/45.7 & 57.4/20.4 & 67.4/28.2 & 71.7/32.3 & 81.5/40.9 & 68.1/37.7 \\

\hdashline

 DW-16-16++ & 0.80 & 44.1/17.1 & 28.5/10.5 & 54.5/28.5 & 83.2/45.6 & 22.4/6.9 & 82.3/38.7 & 55.4/18.9 & 65.2/24.9 & 69.3/28.2 & 76.8/33.0 & 76.0/42.7 \\
 DW-32-16++ & 1.12 & 44.7/17.3 & 25.2/10.0 & 48.8/25.2 & \textbf{79.0}/43.7 & 20.2/5.0 & 76.4/35.4 & \textbf{50.0/15.9} & 60.1/21.8 & \textbf{63.2/24.7} & \textbf{73.5/31.5} & 67.2/38.1 \\ \hdashline

 No-filter \\
 $\quad-$ DW-16-16 & 0.0 & 48.3/19.1 & 34.2/13.0 & 60.2/33.9 & 96.8/61.3 & 24.9/7.9 & 93.8/53.1 & 58.9/22.4 & 72.3/30.2 & 75.5/37.2 & 84.8/42.3 & 83.9/51.6 \\ 
 $\quad-$ DW-32-16 & 0.0 & 47.2/18.8 & 29.0/11.8 & 58.3/35.0 & 92.5/60.8 & 23.5/7.0 & 88.0/64.2 & 55.3/22.4 & 65.5/27.9 & 72.5/38.8 & 80.4/53.8 & 76.7/55.5 \\ 

 \midrule
 \textbf{Ours}\\
 \midrule

% \rowcolor{lightgreen}
 UDW-16-16 \\

% \rowcolor{lightgreen}
 $\quad-$ nll & 0.0 & 49.12/19.55 & 32.18/12.64 & 60.76/32.4 & 94.05/57.88 & 25.75/8.09 & 88.88/44.93 & 70.14/28.53 & 59.84/22.8 & 78.73/38.95 & 93.3/50.73 & 89.2/56.9 \\
% \rowcolor{lightgreen}
 $\quad-$ pesq & 0.0 & 49.46/19.73 & 34.27/13.55 & 60.58/31.99 & 95.63/57.22 & 26.8/8.64 & 87.91/43.83 & 71.92/30.19 & 60.01/22.47 & 76.7/35.72 & 87.51/45.81 & 84.7/52.1 \\

% \rowcolor{lightgreen}
 $\quad-$ entropy & 0.0 & 49.03/19.5 & 33.4/13.57 & 62.74/35.19 & 96.45/58.9 & 25.81/8.29 & 90.31/49.75 & 61.72/25.08 & 71.29/30.42 & 80.69/40.86 & 101.99/55.25 & 96.8/63.7  \\ 

% \rowcolor{lightgreen}
 $\quad-$ conf & 0.0 & 48.73/19.3 & 34.01/13.44 & 59.2/32.17 & 89.77/53.36 & 24.68/7.86 & 89.61/45.69 & 60.05/22.57 & 71.27/28.52 & 76.71/35.08 & 85.44/42.53
 & 84.0/51.3  \\ 

% \rowcolor{lightgreen}
 $\quad-$ proxy & 0.0 & 47.9/18.76 & 30.96/12.7 & 56.03/30.8 & 86.44/52.66 & 24.69/9.25 & 81.7/45.74 & 68.43/26.62 & 58.08/22.07 & 72.96/34.2 & 84.6/43.64 & 74.1/48.2 \\ 

% \rowcolor{lightgreen}
 $\quad-$ sonar & 0.0 & 48.82/19.4 & 33.88/13.19 & 61/33.03 & 87.08/52.03 & 27.75/8.91 & 87.6/45.09 & 62.72/24.81 & 71.92/30.21 & 76.9/35.99 & 82.49/39.72
 & 78.9/47.3 \\

\hdashline

% \rowcolor{lightgreen}
 UDW-32-16 \\
% \rowcolor{lightgreen}
 $\quad-$ nnll & 0.0 & 45.55/17.88 & 29.02/11.52 & 54.66/29.79 & 96.01/57.91 & 20.19/6.13 & 84.03/39.91 & 63.45/23.8 & 52.63/18.43 & 71.98/33.71 & 84.57/43.01 & 79.6/48.0 \\
% \rowcolor{lightgreen}
 $\quad-$ pesq & 0.0 & 47.5/19.0 & 32.0/12.4 & 55.6/30.8 & 88.9/53.8 & 23.5/7.1 & 89.3/44.9 & 55.2/19.6 & 66.4/25.6 & 75.0/34.2 & 83.3/40.4 & 79.0/50.0 \\

% \rowcolor{lightgreen}
 $\quad-$ entropy & 0.0 & 45.68/18.18 & 29.86/12.07 & 53.88/32.97 & 88.92/68.55 & 21.55/6.32 & 91.78/84.74 & 54.32/23.38 & 66.47/33.82 & 73.62/47.04 & 83.91/56.95 & 74.7/54.1  \\

% \rowcolor{lightgreen}
 $\quad-$ conf & 0.0 & 47.07/18.29 & 31.12/11.62 & 54.04/28.14 & 85.53/51.41 & 22.78/6.68 & 81.01/39.84 & 65.69/25 & 69.83/29.36 & 55.32/20.11 & 86.59/44.52 & 72.9/43.6  \\

% \rowcolor{lightgreen}
 $\quad-$ proxy & 0.0 & 45.38/17.71 & 28.22/11.32 & 51.47/27.06 & 84.54/48.86 & 21.41/7.93 & 81.25/37.94 & 53.42/18.05 & 63.9/23.76 & 70.19/29.16 & 78.38/34.03 & 71.0/44.1 \\
% \rowcolor{lightgreen}
 $\quad-$ sonar & 0.0 & 44.74/17.38 & 27.35/10.35 & 52.72/28.08 & 82.33/49.95 & 21.05/6.29 & 80.58/38.85 & 52.29/17.53 & 63.17/22.67 & 67.71/27.87 & 75.42/32.57 & 65.0/37.7 \\

\hdashline

% \rowcolor{lightgreen}
 UDW-16-16++ \\
% \rowcolor{lightgreen}
 $\quad-$ proxy & 0.0 & 52.0/22.5 & 28.4/11.6 & 53.8/28.8 & 86.5/50.6 & 21.9/6.7 & 83.6/43.2 & 54.2/18.8 & 64.9/24.1 & 69.3/30.9 & 78.9/37.2 & 66.2/37.3 \\ 
% \rowcolor{lightgreen}
 $\quad-$ sonar & 0.0 & 45.5/17.7 & 29.9/11.4 & 53.9/28.8 & 83.7/50.1 & 22.8/7.0 & 82.6/40.6 & 55.4/19.6 & 65.0/24.4 & 70.2/29.6 & 77.5/33.9 & 65.7/36.0 \\
% \rowcolor{lightgreen}
 UDW-32-16-++ \\
% \rowcolor{lightgreen}
 $\quad-$ proxy & 0.0 & 44.9/17.5 & 26.6/10.5 & 50.4/25.8 & 84.4/47.1 & 20.7/6.1 & 81.6/39.0 & 51.5/17.0 & 62.6/23.0 & 66.5/26.6 & 77.6/33.8 & 62.1/33.6	  \\ 
% \rowcolor{lightgreen}
 $\quad-$ sonar & 0.0 & 44.2/17.1 & 29.0/11.5 & 51.2/27.0 & 81.7/47.2 & 20.6/5.7 & 79.5/36.5 & 62.3/22.2 & 52.5/17.5 & 67.1/26.8 & 76.0/31.8 & 54.7/31.3 \\

% & DW-1M & & 43.6/16.8 & 26.2/9.9 & 51.6/27.1 & 81.9/44.1 & 20.2/6.2 & 80.5/36.1 & 52.5/17.0 & 62.7/21.9 & 66.2/26.0 & 75.6/31.5 & 56.1/23.7 \\
\bottomrule 
\end{tabular}
}
\caption{ \label{tab:ortho-results-appendix}
WER/CER ($\downarrow$) scores on orthographic transcription. Average is the mean score across all the evaluation sets. All distilled models are trained with a filtering threshold of 80. We report the score on the test split of each dataset. Best results are shown in \textbf{bold}. Second best results are \underline{underlined}. We report the score on the test split of each dataset.
Abbreviations. \textbf{W} - Whisper, \textbf{FT} - Finetuned, \textbf{M} - Medium, \textbf{L} - Large, \textbf{S} - Small, \textbf{DW} - Distill Whisper, \textbf{UDW} - Unsupervised Distill Whisper, \textbf{nll} - negative log likelihood, \textbf{conf} - confidence score.
}
\end{table*}
% \begin{table}[]
% \begin{tabular}{llll}
% \toprule
% Metric     & $\lambda>=80$ & $\lambda >=40$ & $\lambda>=20$ \\
% \midrule
% Conf &             &                &                \\
% SonarSim   &             &                &                \\
% Proxy-ref  &             &                &                \\ \midrule
% \end{tabular}
% \end{table}

% \section{Efficacy}\label{appendix:efficay}
% \input{figures/rocauc}

% This is an appendix.

\end{document}